\documentclass[11pt]{article}
\usepackage{amssymb}
\usepackage{subcaption}
\usepackage{graphicx}
\usepackage{booktabs}
\usepackage{newtxtext}
\usepackage{newtxmath}
\usepackage{latexsym}
\usepackage[a-1b]{pdfx}

\usepackage[final]{acl}

 \usepackage{microtype}

\usepackage[english]{babel} 

\title{SPADE: Structured Prompting Augmentation for Dialogue Enhancement in Machine-Generated Text Detection}

\author{
  \begin{tabular}{cc}
    Haoyi Li* & Angela Yifei Yuan* \\
    The University of Melbourne & The University of Melbourne \\
    \texttt{haoyil4@student.unimelb.edu.au} & \texttt{angela.yuan@student.unimelb.edu.au} \\
    \\
    Soyeon Caren Han$^{\dagger}$ & Christopher Leckie \\
    The University of Melbourne & The University of Melbourne \\
    \texttt{Caren.Han@unimelb.edu.au} & \texttt{caleckie@unimelb.edu.au} \\
  \end{tabular}
}

\begin{document}

\maketitle
\begin{abstract}
The increasing capability of large language models (LLMs) to generate synthetic content has heightened concerns about their misuse, driving the development of Machine-Generated Text (MGT) detection models. However, these detectors face significant challenges due to the lack of high-quality synthetic datasets for training.
To address this issue, we propose SPADE, a structured framework for detecting synthetic dialogues using prompt-based positive and negative samples. Our proposed methods yield 14 new dialogue datasets, which we benchmark against eight MGT detection models. The results demonstrate improved generalization performance when utilizing a mixed dataset produced by proposed augmentation frameworks, offering a practical approach to enhancing LLM application security. Considering that real-world agents lack knowledge of future opponent utterances, we simulate online dialogue detection and examine the relationship between chat history length and detection accuracy. Our open-source datasets, code and prompts can be downloaded from https://github.com/AngieYYF/SPADE-customer-service-dialogue.
\end{abstract}

\makeatletter
\let\@makefnmark\relax
\makeatother
\footnotetext{* Both authors contributed equally to this research.}
\footnotetext{$^{\dagger}$ Corresponding author}

\section{Introduction} \label{introduction}

Large language models (LLMs) are increasingly deployed in conversational systems, but their accessibility also enables adversaries to launch automated attacks. For instance, in online customer service chatrooms, attackers may use LLMs to launch prompt injection attacks that spread misinformation, or flood the system with realistic but excessive queries, leading to denial-of-service outcomes~\cite{61-owasp-2023, Zhan-2024}. These scenarios highlight the pressing need for high-performance Machine-Generated Text (MGT) detection in dialogue settings. However, existing detectors often fail due to the scarcity of high-quality datasets with dynamic dialogue contexts, where traditional data collection methods are time-consuming and expensive, limiting scalability.

Significant research has been conducted on MGT detection~\cite{40-kirchenbauer2024, 39-lu2024,30-bahad-etal-2024, 64-Koike-2024}, focusing on long-form texts such as Reddit~\cite{26-Mitchell-2023}, news articles~\cite{66-Li_2024, 29-Wang-2023}, Wikipedia entries~\cite{Wahle-2022}, and student essays~\cite{64-Koike-2024, Wahle-2022}. However, these types of texts differ fundamentally from dialogues, which are shorter, turn-based, and involve dynamic interactions between two parties that evolve as the conversation progresses. Traditional detection methods, which are designed for static, longer passages, struggle to handle the fluid and interactive nature of dialogues. This challenge is further exacerbated by the lack of high-quality, domain-specific dialogue datasets, which makes it difficult to develop robust MGT detection methods for conversational environments.
The scarcity of suitable dialogue data has been a longstanding issue, and recent methods still have not fully addressed this problem. Collecting real-life dialogues from systems or LLM users is not only expensive but also impractical at scale. To overcome these limitations, data augmentation has emerged as a viable, cost-effective alternative~\cite{43-sennrich-etal-2016, 59-kojima2023, 50-mao2024, 51-labruna2023}. However, maintaining fluency, coherence, and consistency with user goals across multiple interaction stages remains a challenge. Moreover, relying solely on a single augmentation method can limit model generalization, leading to poor performance when encountering out-of-distribution data~\cite{65-Hays_2023}.

In this paper, we propose five novel data augmentation frameworks, specifically designed for synthetic user dialogue generation.
Note that the development of LLM-based chatbot detection models faces several key challenges: (1) scarcity of high-quality training data, (2) high costs associated with collecting real-life dialogue datasets, (3) maintaining coherence and fluency in augmented synthetic dialogues, and (4) inefficiencies in performing detection only after dialogue completion.
Hence, the proposed frameworks employ a structured prompting approach to generate 14 sets of dialogues, which significantly reduce the costs associated with traditional data collection methods. Our frameworks, Partial-Chatbot and Full-Chatbot, are tailored to the interactive and dynamic nature of dialogue. Through automated and manual quality assurance, we ensure that the generated dialogues are fluent and closely aligned with user goals. Additionally, the frameworks support the simulation of online conversational environments, facilitating offline and real-time detection. The datasets are benchmarked against eight MGT detection models, demonstrating improved generalization performance when trained on a mixture of datasets created using our augmentation techniques.
Upon simulating real-world settings, where agents are unaware of future user utterances, we observe a positive correlation between the volume of chat history and detection performance. The proposed datasets and methods enhance MGT detection in dialogues, particularly in cases with limited or incomplete chat history.

The contributions of this paper are: 

\begin{enumerate} 
\item We introduce novel, training-free data augmentation frameworks specifically designed for synthetic user dialogue generation. These frameworks produce 14 new dialogue datasets applicable across various domains, addressing the scarcity of high-quality dialogue data for MGT detection.

\item We refine and enhance domain-specific datasets, ensuring that the dialogues are coherent and aligned with specified user goals, offering a template for other domain-specific applications.

\item We benchmark the performance of these datasets across eight baseline models, demonstrating improved generalizability through the combination of diverse data augmentation methods.

\item We simulate real-time conversations and show that detection accuracy improves as chat history increases, reinforcing the importance of progressive detection. 
\end{enumerate} 

To the best of our knowledge, this is the first work to introduce training-free dialogue data augmentation frameworks applicable to offline and online environments, advancing the detection of MGT across diverse dialogue settings.

\section{Related Work}

\subsection{Dialogue Datasets} 
There are several open-sourced multi-domain dialogue datasets, such as MultiWOZ~\cite{2-eric-2019}, SGD~\cite{3-rastogi-2020}, CrossWOZ~\cite{10-zhu-etal-2020}, and EmoWOZ~\cite{feng-etal-2022-emowoz}, which include customer service dialogues. These datasets only feature dialogues with human users, limiting their effectiveness for detection aimed at identifying synthetic users. While recent work~\cite{6-zheng-2024} has introduced datasets containing LLM-based synthetic users in customer service scenarios, it still falls short of addressing the critical need for extensive dialogue datasets specifically containing synthetic user utterances. 
Our research addresses this limitation by introducing cost-effective and training-free data augmentation frameworks that generate high-quality synthetic user dialogues.


\subsection{Data Augmentation}
Acquiring high-quality labelled training datasets is a costly and challenging task, leading to the development of various data augmentation methods to address data scarcity.  
Paraphrasing was initially conducted in early studies using back translation~\cite{43-sennrich-etal-2016}. With advancements in deep learning, researchers have developed specialized paraphrasing models, such as DIPPER~\cite{49-krishna2023} and BART~\cite{44-lewis-etal-2020, 47-okur2022}. 
Goal-to-dialogue generation creates synthetic dialogues by prompting LLMs to output entire dialogues given user goals and instructions~\cite{51-labruna2023}.
Similarly, end-to-end conversation generation assigns roles to 2 LLMs and asks them to complete dialogues interchangeably~\cite{51-labruna2023, 52-abdullin2024, 55-zahra-2024}. Although these two methods seem easy to implement, they have widely recognized drawbacks. Different LLMs require varying prompt structures, 
and logic and coherence issues can compromise the overall quality of the dialogue.
Our new data augmentation framework overcomes these challenges by maintaining essential conversational features and employing well-structured prompts tailored to widely used LLMs. 

\subsection{MGT Detection}
MGT detection is a classification task where a model aims to classify a given text into categories such as human versus any subsets of LLMs. Detection approaches can generally be divided into three categories: statistical methods, fine-tuned pretrained models, and feature-based methods.
Statistical methods rely on the different distributions of word choices between humans and language models. Some statistics and proposed models include cross-perplexity ~\cite{binoculars}, entropy ~\cite{24-Lavergne-2008, 25-gehrmann2019}, and log probability~\cite{26-Mitchell-2023, 27-bao2024}, which had outstanding performance in zero-shot MGT detection. Fine-tuning pre-trained transformer models such as BERT and RoBERTa~\cite{29-Wang-2023, 30-bahad-etal-2024, 31-guo2023}, which study semantic features for MGT classification tasks, also have impressive performance. Feature-based models rely on difference in semantic and lexical features between human-written text and MGT~\cite{36-Mindner-2023}. The extracted features serve as input to common machine learning (ML) models for classification.
We have chosen to evaluate our datasets using statistical-based, feature-based, and pretrained LLM-based methods, in order to compare the performance of detectors across different data augmentation frameworks and to evaluate the ability of data augmentation methods to enhance model generalization.

\section{Data Augmentation Framework} \label{DataAugmentation}

\begin{figure*}[t!]
  \centering
  \begin{subfigure}[t]{0.377\textwidth}
    \centering
    \includegraphics[scale = 0.215]{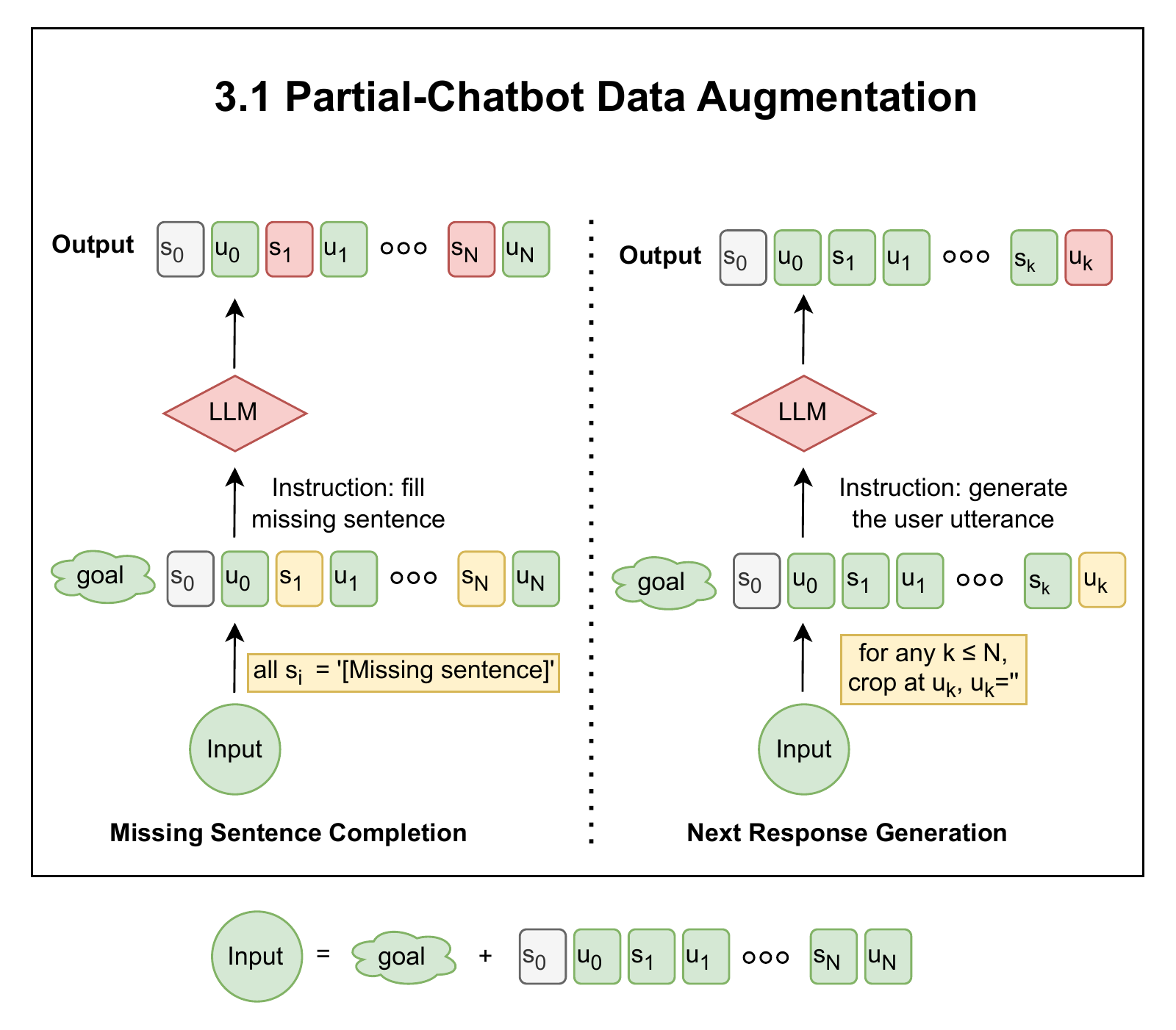}
    \caption{Partial-Chatbot Data Augmentation}
    \label{fig:partial-framework}
  \end{subfigure}
  ~
  \begin{subfigure}[t]{0.6\textwidth}
    \centering
    \includegraphics[scale = 0.215]{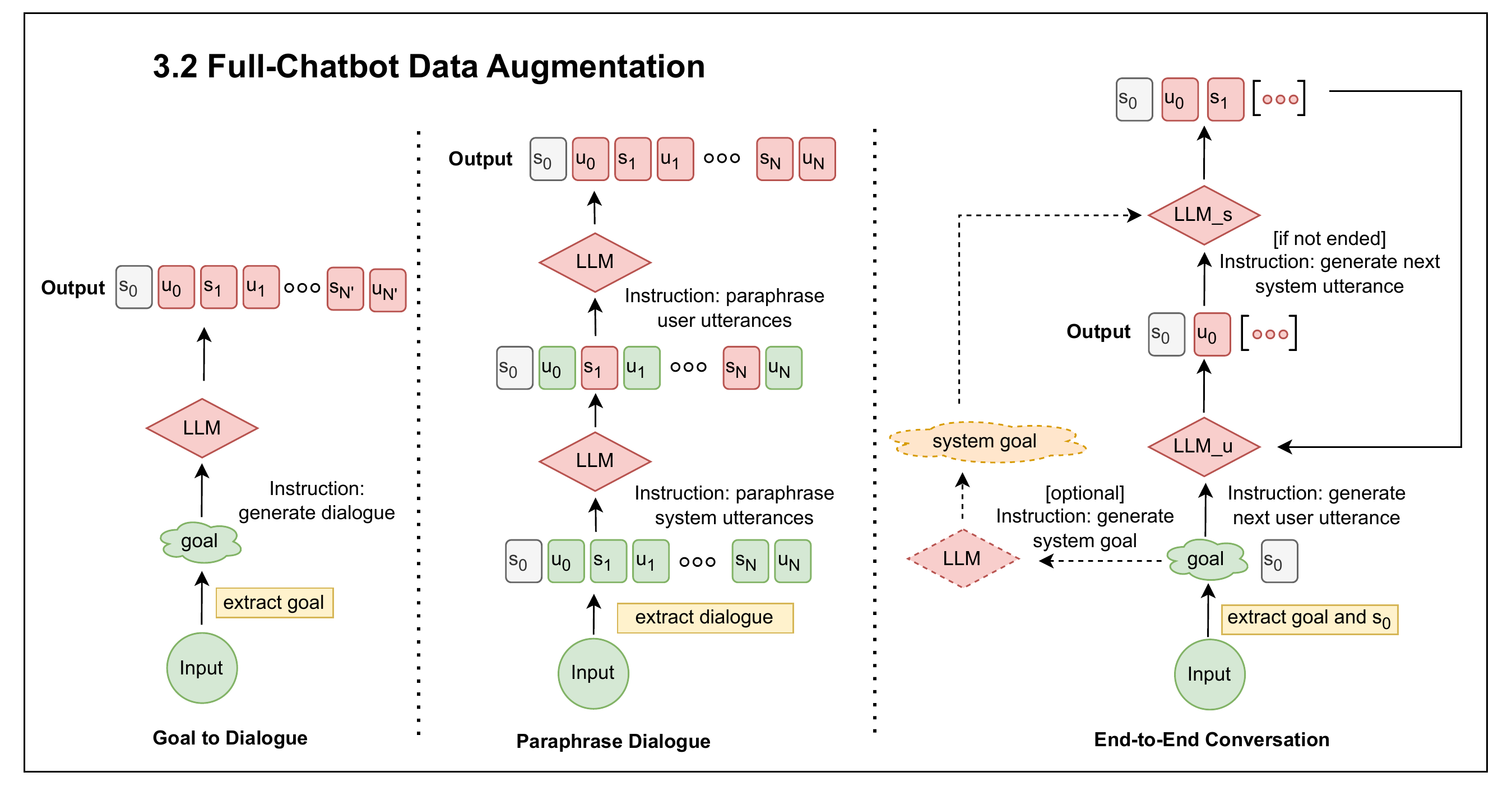}
    \caption{Full-Chatbot Data Augmentation}
    \label{fig:full-framework}
  \end{subfigure}
  \caption{Data augmentation frameworks, where the input is a dialogue alternating between a customer service system ($s_i$) and a human user ($u_i$) with a user goal ($goal$), and outputs a Partial-Chatbot or a Full-Chatbot dialogue. The initial system response $s_0$ can be omitted or set to a standard starting line common to all dialogues. LLM instructions shown here are incomplete.}
  \label{fig:augmentation}
\end{figure*}

This section details the proposed training-free dialogue data augmentation frameworks designed to generate high-quality synthetic dialogues, from bona fide human-generated dialogues. 
These frameworks fall into two main categories: Partial-Chatbot Data Augmentation and Full-Chatbot Data Augmentation.

Figure~\ref{fig:augmentation} outlines the abstract construction process of each framework. Appendix~\ref{prompt example structure} provides example dialogues and complete prompts.

\subsection{Partial-Chatbot Data Augmentation}
The Partial-Chatbot Data Augmentation frameworks generate dialogues partially authored by LLMs, while the remaining dialogue segments retain human-generated utterances. 

\textbf{Missing Sentence Completion: }\label{MissingSentence}
The Missing Sentence Completion, denoted as $f_{MS}$, generates Partial-Chatbot dialogues by filling in the missing sentences for one of the participants in the conversation. 
All system utterances in bona fide dialogues are replaced by synthetic text, to be used as negative samples when positive samples contain synthetic system utterances. This controls the consistency of whether the system is a chatbot or human across positive and negative samples with synthetic and human users respectively. 
For each original dialogue $d_i$, $f_{MS}(d_i) = L(q(d_{i}), t_{MS})$, where $q(d_i)$ replaces all system utterances $d_{ij}^{(s)}$ with ``[missing sentence]", and $t_{MS}$ is a prompt engineered for LLM $L$ to replace missing sentences.
A comprehensive structure for prompt $t_{MS}$ is provided in Appendix \ref{prompt example structure}.

\textbf{Next Response Generation: } The Next Response Generation framework, denoted as $f_R$, produces user responses based on incomplete dialogue history, ensuring consistency by maintaining original system utterances. This framework 
only generates user responses, which serve as positive samples for MGT detection. $f_{R}(d_i) = \{L(\phi(d_i^{k}), g_i, t_R)|k\leq N_i\}$ where the original dialogue $d_i$ with $N_i$ turns is cropped to produce incomplete chat histories $d_i^{k}$ with exactly $k$ turns, and $\phi(d_i^{k})$ replaces the last user utterance $d_{ik}^{(u)}$ with an empty string. $t_R$ is a prompt engineered for $L$ to generate the user response $d_{ik}'^{(u)}$ according to the original goal $g_i$. This approach not only ensures dialogue coherence and goal alignment but also eliminates any reliance on synthetic system responses for detection, thereby enhancing applicability.


\subsection{Full-Chatbot Data Augmentation}
The Full-Chatbot Data Augmentation frameworks generate dialogues in which LLMs produce both system and user utterances. 

\textbf{Goal to Dialogue (G2D):} The G2D framework, denoted as $f_G(g_i) = L(g_i, t_G)$, generates a Full-Chatbot dialogue based on a user goal $g_i$. Unlike traditional few-shot learning methods that require the selection of demonstrations for each goal, G2D employs a structured prompt $t_G$ with comprehensive instructions for the LLM. This approach reduces input token overhead and achieves high goal-dialogue alignment without external training. The prompt $t_G$ is tailored according to the following components: 1) Task Summary: A brief description of the dialogue objective. 2) Example Dialogue: A sample conversation demonstrating the expected interaction flow. 3) Goal-Specific Instructions for User: Detailed guidance on how the synthetic user should respond based on $g_i$. 4) Slot and Domain Knowledge for System: Contextual information is required for the system to provide coherent responses. 5) Conversation Termination Conditions: Criteria for ending the dialogue to ensure it remains goal-oriented. 6) Sensitive Information Masking: Instructions to anonymize sensitive details, such as replacing the exact reference number “AX12387” with “[ref]”.
This structured prompting enables the generation of realistic dialogues that align closely with the user's goal, increasing the diversity and variability of synthetic samples. Dialogues generated using G2D can serve as positive samples compared to those produced by the Missing Sentence Completion framework. Complete prompting templates are provided in Appendix \ref{prompt example structure}.

\textbf{Paraphrase Dialogue (Par.):} The Par. framework ($f_P$) uses an iterative paraphrasing strategy to convert Full-Human dialogues into Full-Chatbot dialogues while preserving the conversational structure and logical flow. The process involves two stages: (i) $d_i^1 = L_{s}(d_i, t_{P, s})$ is a dialogue with all system utterances $d_{ij}^{(s)}$ replaced, and (ii) $d_i^2 = L_{u}(d_i^1, t_{P, u})$ has all user utterances $d_{ij}^{1(u)}$ in stage 1 output replaced. This two-stage approach produces two distinct dialogue sets: $d_i^1$ as negative samples (synthetic system responses only) and $d_i^2$ as positive samples (fully synthetic dialogues). While this method offers limited flexibility in user simulation due to its dependence on the original dialogue’s structure, it enhances the cohesiveness of the system’s utterances. Example prompting structures are in Appendix \ref{prompt example structure}.

\textbf{End-to-End Conversation (E2E Convo.)}: The E2E Convo. framework ($f_L$) generates fully synthetic dialogues by assigning distinct roles (system and user) to two instances of LLMs. The LLMs interact to create a complete dialogue. The prompt structure for E2E Convo. includes: 1) Task Summary: Overview of the dialogue scenario and expected outcomes. 2) Example Dialogue: A sample conversation to illustrate the interaction. 3) Role-Specific Instructions: Detailed guidance for both user and system LLMs. 4) Conversation Termination Conditions: Specifications for when to conclude the interaction. 5) Sensitive Information Masking: Instructions to mask identifiable information, such as replacing “AX12387” with “[ref]”. 6) Chat History Context: Previously exchanged dialogue turns to maintain context. 
The generated dialogues can serve as positive samples for training detection models. In contrast, dialogues generated by the Missing Sentence Completion framework can serve as negative samples. The example prompting structure can be found in Appendix \ref{prompt example structure}.

\section{Dialogue Data Construction}
This section outlines the preprocessing of the Full-Human dataset, synthetic data generation, and quality assurance.

\subsection{Data Source}\label{Fine-tuned MultiWOZ2.1 dataset}
The MultiWOZ 2.1 dataset~\cite{2-eric-2019} contains customer service dialogues like hotel booking, collected using a Wizard-of-Oz setup where two participants act as the user and system. We use this dataset with Covlab3~\cite{5-zhu-2023} labeled user goals as our baseline for applying data augmentation frameworks.
However, goal-dialogue mismatches led to repetitive responses, such as repeatedly asking, ``Does this hotel have WiFi?" across different contexts. This was due to discrepancies between the dialogues and their annotated goals, including missing or incorrect goals. To resolve this, we conducted a two-step refinement (i) Llama 70B~\cite{touvron2023llama} automatically verified goal achievement, and (ii) we manually revised goals to ensure alignment without changing dialogue content. 
Incomplete dialogues were removed, resulting in a final set of 616 out of 623 refined hotel dialogues.

\subsection{Data Collection} \label{Data Collection}
We employ two widely used LLMs to generate the synthetic user datasets, GPT-3.5 ~\cite{openai2023gpt35} and Llama 70B ~\cite{touvron2023llama}. We executed the framework defined in Section \ref{DataAugmentation} to generate Partial-Chatbot and Full-Chatbot synthetic dialogues, utilizing the fine-tuned MultiWOZ 2.1 dataset defined in Section \ref{Fine-tuned MultiWOZ2.1 dataset}. As shown in Table \ref{tab:Dataset Collection}, 14 new datasets are created according to our training-free data augmentation frameworks. We produced 6 Partial- and 8 Full-Chatbot dialogue datasets. 

During the dialogue generation process, we found that LLM-generated dialogues include errors such as meaningless information, dialogues in the wrong format, and repeated utterances. To eliminate these errors, we regenerate the erroneous dialogues until we obtain correct results. The regeneration takes 15 rounds on average for each data augmentation method. 
The entire generation process for all 14 dialogue datasets cost approximately 10 AUD using the API.
The quality of generated responses are assessed using both automatic and manual metrics. We automated the validation of several structural aspects (e.g., interleaving of user-system turns, absence of repetition) and manually review content quality. Re-generation stops when outputs pass these checks.
To further test the robustness of our prompts, we conduct an exchanged prompts experiment for the employed LLMs, detailed in Appendix~\ref{Prompt Exchange Experiment}.

\begin{table}[t]
\caption{14 new datasets generated using different data augmentation frameworks proposed.}
\label{tab:dataset-collection}
\centering
\small
\label{tab:Dataset Collection}
\resizebox{\columnwidth}{!}{%
\begin{tabular}{@{}ccl@{}}
\hline
\textbf{No.} & \textbf{Category} & \textbf{Dataset} \\
\hline
1  & Full-Chatbot     & GPT Par. Full-Chatbot \\
2  &                  & Llama Par. Full-Chatbot \\
3  &                  & GPT G2D \\
4  &                  & Llama G2D \\
5  &                  & GPT-GPT E2E Convo. \\
6  &                  & Llama-Llama E2E Convo. \\
7  &                  & GPT-Llama E2E Convo. \\
8  &                  & Llama-GPT E2E Convo. \\
\hline
9  & Partial-Chatbot  & GPT Par. Chatbot-Human \\
10 &                  & Llama Par. Chatbot-Human \\
11 &                  & GPT Missing Sentence Completion \\
12 &                  & Llama Missing Sentence Completion \\
13 &                  & GPT Next Response Generation \\
14 &                  & Llama Next Response Generation \\
\hline
\end{tabular}
}
\end{table}

\subsection{Quality Assurance} \label{Quality Assurance}
As the flexibility given to LLMs may cause coherence and fluency issues or misalignment between the goal and the augmented synthetic dialogues.
We evaluate the quality of our dataset. For Partial-Chatbot synthetic dialogue, measurements focus on the coherence and fluency of the generated responses about the given chat history. For Full-Chatbot synthetic dialogues, after evaluating the dialogues according to the dimensions mentioned above, we additionally conduct automated and manual quality assurance on the degree of matching between the dialogues and the original goal to further assess the quality of the generated dialogues against the mismatch issues present in Section \ref{Fine-tuned MultiWOZ2.1 dataset}. 

To conduct automated quality assurance on Partial-Chatbot and Full-Chatbot dialogues, we use a pre-trained model called UniEval-dialog~\cite{72-zhong2022}, which measures the generated dialogue responses regarding naturalness, coherence, engagement, groundedness, and understandability. The model outputs `yes' or `no.' We calculate the `yes' rate to illustrate the quality of our generated dialogues.


\begin{table}
    \caption{UniEval-Dialog quality assurance results of generated datasets.}
  \label{tab:Partial-Chatbot data evaluation}
  \centering
  \small
  \resizebox{\columnwidth}{!}{%
    \begin{tabular}{lcc}
        \hline
        \textbf{Dataset} & \textbf{GPT-3.5}& \textbf{Llama 70B}\\
        \hline
        Missing Sentence Completion & 97.24\%& 98.88\%\\
        Next Response Generation & 95.79\% & 96.74\%\\
        Par. Chatbot-Human & 98.55\%& 97.43\%\\
        Par. Full-Chatbot & 99.84\% & 99.52\%\\
        G2D & 100.0\% & 97.73\%\\
        E2E Convo. & 99.75\% & 99.12\%\\
        \hline
    \end{tabular}
    }
\end{table}

Table~\ref{tab:Partial-Chatbot data evaluation} shows how often the generated dialogues were rated as coherent and fluent using the UniEval-dialog model.
All generated dialogue datasets achieved scores above 95\%, demonstrating consistently high quality. The dialogues generated by the three Full-Chatbot data augmentation frameworks achieved scores over 97\%, showing a higher quality compared to Partial-Chatbot dialogues.

To ensure each Full-Chatbot dialogue matches the provided goal, we conduct automatic and human survey-based degrees of match experiment. The results show that our generated dialogues achieve an average goal dialogue match score similar to the original human dialogues. Details of the experiments can be found in Appendix \ref{full chatbot quality assurance}. 


In summary, all synthetic dialogue datasets generated from our training-free data augmentation frameworks have reached a standard that resolves the challenge for dialogue augmentation mentioned in Section \ref{introduction}.

\section{Offline Dialogue Detection} \label{Full Dialogue Detection}

\begin{table*}
    \centering
    \small
    \caption{Macro-F1 of detection models. The highest score of each detection task is in bold.}
    \resizebox{\textwidth}{!}{%
    \begin{tabular}{llcccccccc}
        \hline
         & & \multicolumn{2}{c}{\textbf{Statistical}} & \multicolumn{1}{c}{\textbf{PLM}} & \multicolumn{5}{c}{\textbf{Feature}} \\
        \hline
            \textbf{Dataset} & \textbf{Detection} & \textbf{Entropy} & \textbf{Binoculars} & \textbf{RoBERTa} & \textbf{MLP} & \textbf{XGboost} & \textbf{LogR} & \textbf{SVM} & \textbf{RF}\\
            \hline
            Par. & Binary & 0.6740 & 0.7451 & 0.8942 & 0.9510 & 0.8754 & \textbf{0.9592} & 0.9558 & 0.9455\\
            \hline
            G2D & Binary & 0.6617 & 0.8227 & 0.9913 & \textbf{0.9914} & 0.9432 & 0.9857 & 0.9878 & 0.9699\\
            \hline
            E2E Convo. & Binary & 0.8846 & 0.8227 & 0.9939 & \textbf{0.9976} & 0.9816 & 0.9949 & 0.9949 & 0.9899\\
            \hline
            Next Response Generation & Binary & 0.7176 & 0.6342 & 0.9372 & 0.9414 & 0.8780 & \textbf{0.9537} & 0.9436 & 0.9384\\
        \hline
    \end{tabular}
    }
    \label{tab:performance}
\end{table*}

This section focuses on offline dialogue detection, where a single-stage MGT detection model is applied to all user responses in a dialogue simultaneously. We evaluated the performance of our augmented datasets using eight models for MGT detection.


\subsection{Feature-based Detection}\label{feature based}
Based on the features summarized in~\cite{36-Mindner-2023}, we implemented a selection of features relevant to dialogues, along with utterance counts and derived metrics, including 7 categories: Sentiment, Errors, Readability, Statistic, List Lookup, Document, Text Vector, Derived Features.
Feature vectors with 910 dimensions were extracted from user utterances only. After scaling and F-test feature selection, these vectors were used as inputs for traditional ML models like Logistic Regression (LogR)~\cite{Joseph-1944}, Support Vector Machine (SVM)~\cite{SVM-Cristianini2008}, Random Forest (RF)~\cite{RF-breiman2001random}, Multilayer Perceptron (MLP)~\cite{MLP-Taud2018}, and XGBoost~\cite{38-chen-2016}.

\subsection{Statistical-based Detection}
Inspired by the entropy-based MGT detection method from Liu~\cite{25-gehrmann2019}, for each dialogue $d = \{w_0, \cdots, w_n\}$ consisting of $n$ word tokens across  $n_{sen}$ user utterances, we calculate entropy as follows: $$S = -\frac{\Sigma_{i=0}^n P(w_i) log_2 (P(w_i))}{n_{sen}}$$ 
$P(w_i) = \frac{n_{i}}{N}$, where $n_i$ represents the frequency of word $w_i$ in the dataset, and $N$ is the total number of tokens in the dataset. 
In addition to entropy, we include Binoculars~\cite{binoculars} as another statistical model based on cross-perplexity allowing evaluation without task-specific fine-tuning.

\subsection{PLM-based Detection}
We adopted the PLM (Pretrained Language Model)-based detection model from Seq-RoBERTa~\cite{29-Wang-2023}, where dialogue-level predictions are made using hard voting across token-level predictions. Padding labels were applied to non-user tokens to exclude them from model training and final prediction. For the Next Response Generation dataset, meaningful labels were assigned only to the last user response, with all other tokens using padding labels.

\subsection{Experiments and Results} \label{Experiments and Results}
We conducted experiments using all eight models from three detection method categories, testing across 14 datasets each containing 616 dialogues, generated using Partial-Chatbot and Full-Chatbot data augmentation frameworks (Section \ref{DataAugmentation}). The models were evaluated on both multiclass classification (distinguishing ``Human,``GPT", and ``Llama") and binary classification (``human" and ``AI"). The evaluation metric used was the macro F1 score, where low scores indicate misclassifications for both chatbot and human detection.
For the Next Response Generation dataset, each human response served as a negative sample, and each generated next response was used as a positive sample. For full dialogue datasets, negative and positive samples were defined as per Section \ref{DataAugmentation}. We randomly split 80\% of each dataset by dialogue ID for training and reserved 20\% for testing. Each model was trained and tested five times, and the mean F1 score was reported.

Table~\ref{tab:performance} presents the detailed results of our experiments for binary classification, and the results for multiclass classification can be found in Appendix~\ref{sec:multiclass classification}. Several key observations can be drawn. First, the detection performance on the Par. dataset was consistently lower across all models compared to the G2D and E2E Convo. datasets. This suggests that paraphrasing introduces greater complexity for MGT detection, making it harder for models to identify synthetic dialogues. For instance, in binary classification, the MLP model achieved an F1 score of 0.95 on the Par. dataset, but a perfect score of 0.99 for both the G2D and E2E Convo. datasets. This performance gap indicates that paraphrased dialogues, by altering sentence structures while maintaining semantic meaning, are more difficult for models to detect as machine-generated, unlike the synthetic dialogues generated by G2D or E2E Convo., where the structure remains more predictable.
Second, datasets containing single sentences, like Next Response Generation, exhibited consistently lower performance compared to full dialogue datasets. This highlights the importance of longer context in improving detection accuracy, as models can leverage richer linguistic features and patterns. For example, the MLP model's F1 score for Next Response Generation was 0.9414, significantly lower than the 0.99 achieved for both G2D and E2E Convo. datasets. This suggests that with more conversational turns and context, models can better differentiate between human and AI-generated dialogues by identifying language and interaction flow inconsistencies.
Based on these findings, the best-performing models for our task are PLM-based and feature-based methods, especially RoBERTa, MLP, and LogR, which are employed in subsequent experiments to further explore MGT detection performance.

\subsection{Cross Dataset Detection}\label{sec:Cross-Dataset-Detection}

\begin{figure}[t]
  \centering
  \includegraphics[width=\linewidth]{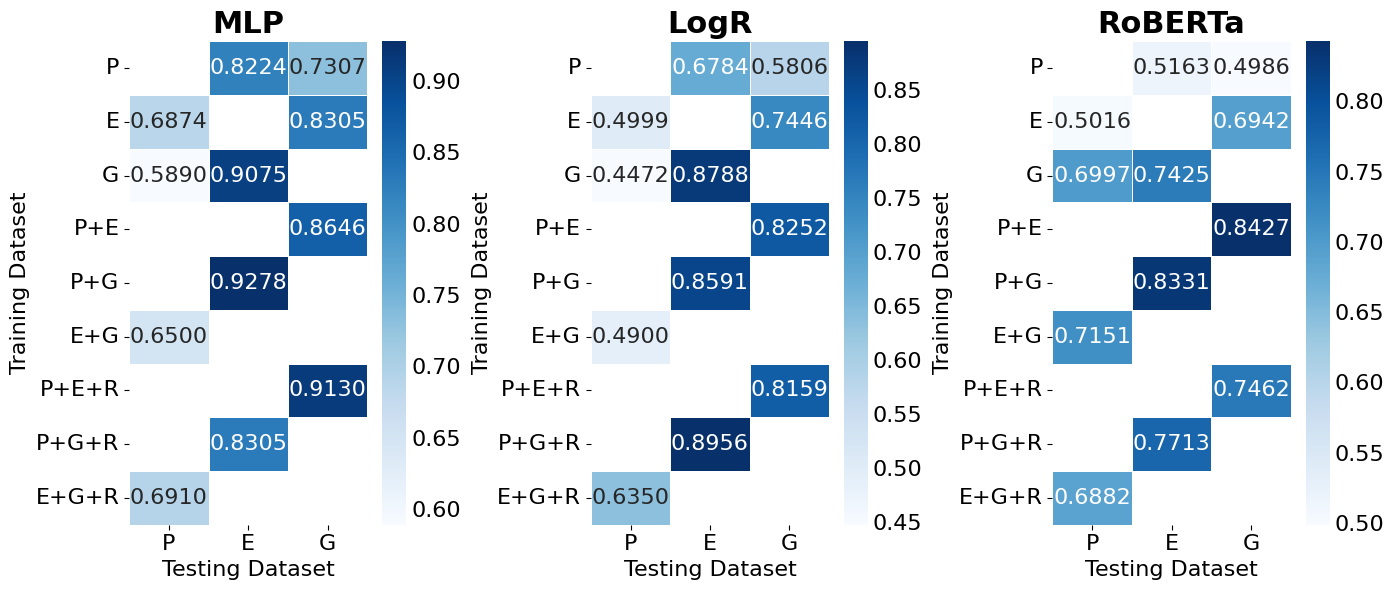}
  \caption{Performance for Cross Dataset Detection with LogR, MLP, and RoBERTa trained on the column-indexed dataset and tested on the row-indexed dataset. $P$ represent Par. dataset, $E$ represent E2E Convo. dataset, $G$ represent G2D dataset, and $R$ represent Next Response Generation dataset.}
  \label{heatmap cross dataset}
\end{figure}

To evaluate generalizability, we tested the top three models (MLP, LogR, and RoBERTa), on cross-dataset binary classification between human and AI-generated dialogues. We use Par., G2D and E2E Convo. datasets separately as testing datasets. To form a training dataset, we randomly sample 1246 dialogues with equally distributed human and synthetic data from different combinations of datasets other than the testing dataset. For RoBERTa, we fine-tuned the model for four epochs on each training dataset. For feature-based models MLP and LogR, we use the trained vectorizer from the training dataset to derive the TFIDF features for the dialogues in the testing dataset.


Figure \ref{heatmap cross dataset} depicts the F1 scores for 12 different combinations of training and testing datasets, demonstrating that combining datasets from various data augmentation frameworks improves model generalizability. MLP models trained solely on the Par. dataset achieved an F1 score of 0.7307 on the G2D test set, whereas MLPs trained on both the Par. and E2E Convo. datasets performed significantly better, achieving an F1 score of 0.8646. Both models were trained on the same number of samples drawn from different sources. This underscores that training with aggregated datasets from diverse augmentation methods results in better generalization to out-of-distribution data.
However, combining complete dialogues with single utterances in training datasets does not continually improve performance. When the Next Response Generation dataset was added to the training set, RoBERTa's performance dropped from above 0.80 to below 0.80—when tested on E2E Convo. and G2D datasets. This suggests that mixing data types can introduce noise and hinder the model's ability to detect complete synthetic dialogues.
To provide further proof our conclusion on generalisability, we conducted additional experiments on other out-of-distribution datasets and different generators. Details about the setup of the experiment and the results can be found in the Appendix \ref{appendix:external-data}.

\section{Online Dialogue Detection} 
\begin{table}[t]
    \centering
    \small
    \caption{Macro-F1 on exact number of utterances.}
    \resizebox{\columnwidth}{!}{%
    \begin{tabular}{llccccccc}
    \hline
            \textbf{Dataset} & \textbf{\#Utterance} & \textbf{MLP} & \textbf{LogR} & \textbf{RoBERTa} \\
            \hline
            E2E Convo. & 1 & 0.7446 & \textbf{0.9063} & 0.7095 \\
            \textbf{} & 2 & 0.8142 &0.9538 & \textbf{0.9545} \\
            \textbf{} & 3 & 0.8807 &0.9683 & \textbf{0.9825} \\
            \textbf{} & 4 & 0.8828 &0.9809 & \textbf{0.9949}\\
            \hline
            G2D. & 1 & \textbf{0.8505} & 0.8080 & 0.7560 \\
            \textbf{} & 2 &  0.8779 & 0.8386 & \textbf{0.9012} \\
            \textbf{} & 3 & \textbf{0.9301} & 0.9031 & 0.9228 \\
            \textbf{} & 4 &  \textbf{0.9618} & 0.9543 & 0.9456 \\
            \hline
            Par. & 1 & \textbf{0.6303} & 0.6070 & 0.6293 \\
            \textbf{} & 2 & 0.7525 & \textbf{0.7528} & 0.7174 \\
            \textbf{} & 3 & 0.7836 & \textbf{0.8504} & 0.7969 \\
            \textbf{} & 4 & 0.8013 & \textbf{0.8767} & 0.8356 \\
        \hline
    \end{tabular}
    }
    \label{tab:progressing_performance}
\end{table}

\begin{figure}[t]
  \centering
  \includegraphics[width=\linewidth]{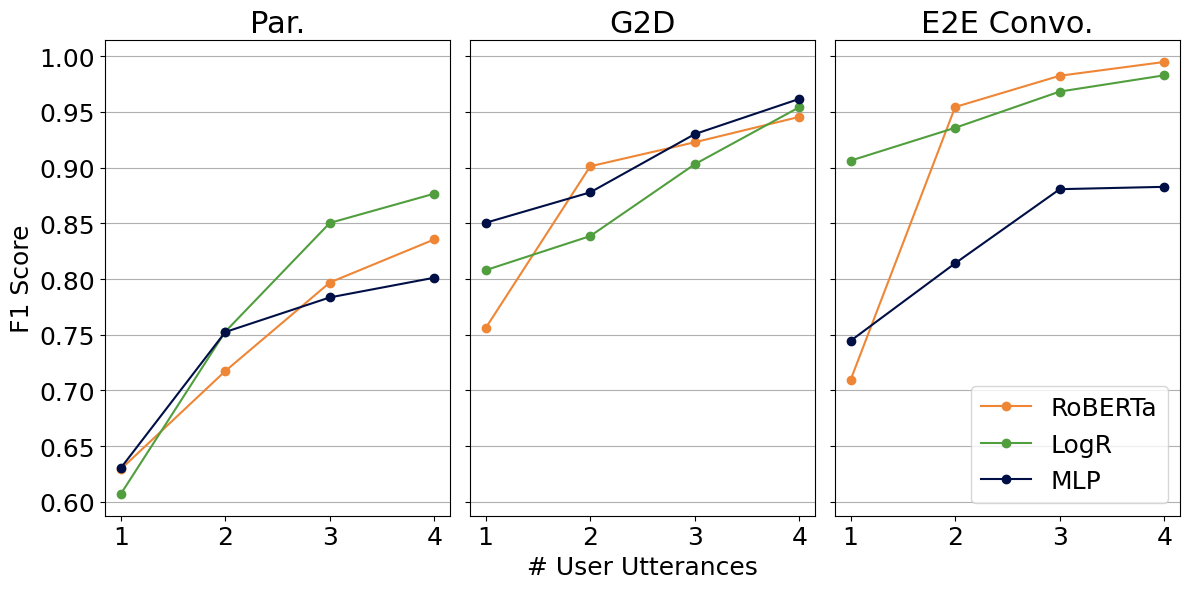}
  \caption{The line graph illustrates a positive relationship between detection rate and the number of user utterances.}
  \label{fig:progressing_performance}
\end{figure}

In an online dialogue setting, both the customer service system and the user can access the chat history but lack knowledge of future utterances. This dynamic introduces a new challenge: detecting chatbot users based on incomplete chat history, with updates occurring as new user utterances are progressively received. Unlike common MGT detection, which receives a passage as input, a new challenge is to detect the chatbot user based on the incomplete chat history and update the detection as user utterances are progressively received. The proposed E2E Convo. framework addresses this challenge by emulating an online environment with turn-based interaction between a system and a user. To conduct online dialogue detection on the generated dataset, we use the high-performing models in the offline dialogue detection setting (Section \ref{Experiments and Results}, including MLP, LogR, and RobERTa). 

Experiments were conducted on the E2E Convo., Par., and G2D datasets to explore performance trends across different prompting structures. Models were initially trained on complete dialogues, and during detection, the input dialogues were progressively cropped to varying lengths while maintaining the same binary classification labels as the full dialogues.  We tested the models on dialogues cropped to $k \in \{1,2,3,4\}$ user utterances, as dialogues with less than k user utterances are excluded from testing datasets, and many dialogues terminate after 4 utterances.

On both feature-based models (MLP and LogR) and a PLM-based model (RoBERTa), Figure \ref{fig:progressing_performance} reveals a positive correlation between detection accuracy and the number of user utterances across all datasets. As shown in Table \ref{tab:progressing_performance}, on the E2E Convo. dataset, MLP and RoBERTa showed F1 scores below 0.8 when only the first user utterance was available. In contrast, LogR achieved an F1 score of 0.9063 with a single utterance, which increased to 0.9809 when four utterances were available. This performance trend highlights the reduced detection accuracy with limited context, which delays the identification of AI users and increases resource consumption. Compared to full dialogue detection (Section \ref{Full Dialogue Detection}), the reduced accuracy in online detection with partial conversations underscores the need for future work on continuous chatbot detection. Two methods to enhance the detection of incomplete dialogues are presented in Appendix \ref{appendix:online_dia}, and a qualitative analysis of online detection performance is provided in Appendix \ref{qualitative online dialogue detection}.



\section{Conclusion}
The scarcity of high-quality datasets for MGT detection in customer service remains challenging. This paper introduced five structured data augmentation frameworks to reduce the costs of traditional dialogue collection, using prompting techniques to efficiently generate synthetic dialogues. We derived 14 new datasets and evaluating these across eight MGT detection models, we found that training on a mix of these datasets significantly improves generalization. Simulated online dialogue detection showed that longer chat histories enhance detection accuracy. Given the rising threats of misused LLMs in real-world systems, SPADE contributes a practical and extensible methodology for improving LLM safety.

\section*{Limitations}
We present our data augmentation framework, which generates new datasets focused on customer service dialogues. A significant challenge is enabling continuous learning for the detector, as new utterances are incrementally introduced. Retraining the model every time when a new utterance is received is inefficient due to resource constraints. Our experiments revealed that expanding datasets with varied chat history scenarios can effectively improve the early detection capabilities of current models. This approach helps the detector generalize across different interaction patterns and user behaviors. However, continuous learning strategies that allow models to update automatically to new utterances without the need for full retraining are an important area for future research. We believe that exploring these strategies will further enhance the efficiency and scalability of synthetic dialogue detection systems.

\section*{Ethics Statement}
We follow the Code of Ethics. No private data or non-public information is used in this work. For manual quality assurance, we invited 15 volunteers, including 5 bachelor students, 8 master students and 2 PhD students. The human survey does not involve any personally sensitive information.

\section*{Acknowledgments}
This work was supported in part by the Australian Research Council Centre of Excellence for Automated Decision-Making and Society, and by the Australian Internet Observatory, which is co-funded by the Australian Research Data Commons (ARDC) through the HASS and Indigenous Research Data Commons.

\bibliography{spade}

\begin{thebibliography}{42}
\providecommand{\natexlab}[1]{#1}

\bibitem[{Abbasiantaeb et~al.(2024)Abbasiantaeb, Yuan, Kanoulas, and Aliannejadi}]{55-zahra-2024}
Zahra Abbasiantaeb, Yifei Yuan, Evangelos Kanoulas, and Mohammad Aliannejadi. 2024.
\newblock \href {https://doi.org/10.1145/3616855.3635856} {Let the llms talk: Simulating human-to-human conversational qa via zero-shot llm-to-llm interactions}.
\newblock In \emph{Proceedings of the 17th ACM International Conference on Web Search and Data Mining}, WSDM '24, page 8–17, New York, NY, USA. Association for Computing Machinery.

\bibitem[{Abdullin et~al.(2023)Abdullin, Molla, Ofoghi, Yearwood, and Li}]{52-abdullin2024}
Yelaman Abdullin, Diego Molla, Bahadorreza Ofoghi, John Yearwood, and Qingyang Li. 2023.
\newblock \href {https://aclanthology.org/2023.gem-1.16} {Synthetic dialogue dataset generation using {LLM} agents}.
\newblock In \emph{Proceedings of the Third Workshop on Natural Language Generation, Evaluation, and Metrics (GEM)}, pages 181--191, Singapore. Association for Computational Linguistics.

\bibitem[{Bahad et~al.(2024)Bahad, Bhaskar, and Krishnamurthy}]{30-bahad-etal-2024}
Sankalp Bahad, Yash Bhaskar, and Parameswari Krishnamurthy. 2024.
\newblock \href {https://doi.org/10.18653/v1/2024.semeval-1.132} {Fine-tuning language models for {AI} vs human generated text detection}.
\newblock In \emph{Proceedings of the 18th International Workshop on Semantic Evaluation (SemEval-2024)}, pages 918--921, Mexico City, Mexico. Association for Computational Linguistics.

\bibitem[{Bao et~al.()Bao, Zhao, Teng, Yang, and Zhang}]{27-bao2024}
Guangsheng Bao, Yanbin Zhao, Zhiyang Teng, Linyi Yang, and Yue Zhang.
\newblock Fast-detectgpt: Efficient zero-shot detection of machine-generated text via conditional probability curvature.
\newblock In \emph{The Twelfth International Conference on Learning Representations}.

\bibitem[{Berkson(1944)}]{Joseph-1944}
Joseph Berkson. 1944.
\newblock Application of the logistic function to bio-assay.
\newblock \emph{Journal of the American statistical association}, 39(227):357--365.

\bibitem[{Breiman(2001)}]{RF-breiman2001random}
Leo Breiman. 2001.
\newblock \href {https://doi.org/10.1023/A:1010933404324} {Random forests}.
\newblock \emph{Machine Learning}, 45(1):5--32.

\bibitem[{Chen and Guestrin(2016)}]{38-chen-2016}
Tianqi Chen and Carlos Guestrin. 2016.
\newblock \href {https://doi.org/10.1145/2939672.2939785} {Xgboost: A scalable tree boosting system}.
\newblock In \emph{Proceedings of the 22nd ACM SIGKDD International Conference on Knowledge Discovery and Data Mining}, KDD '16, page 785–794, New York, NY, USA. Association for Computing Machinery.

\bibitem[{Cristianini and Ricci(2008)}]{SVM-Cristianini2008}
Nello Cristianini and Elisa Ricci. 2008.
\newblock \href {https://doi.org/10.1007/978-0-387-30162-4_415} {\emph{Support Vector Machines}}, pages 928--932.
\newblock Springer US, Boston, MA.

\bibitem[{Eric et~al.(2020)Eric, Goel, Paul, Kumar, Sethi, Goyal, Ku, Agarwal, Gao, and Hakkani-T{\"u}r}]{2-eric-2019}
Mihail Eric, Rahul Goel, Shachi Paul, Adarsh Kumar, Abhishek Sethi, Anuj~Kumar Goyal, Peter Ku, Sanchit Agarwal, Shuyang Gao, and Dilek Hakkani-T{\"u}r. 2020.
\newblock Multiwoz 2.1: A consolidated multi-domain dialogue dataset with state corrections and state tracking baselines.
\newblock In \emph{12th International Conference on Language Resources and Evaluation, LREC 2020}, pages 422--428. European Language Resources Association (ELRA).

\bibitem[{Feng et~al.(2022)Feng, Lubis, Geishauser, Lin, Heck, van Niekerk, and Gasic}]{feng-etal-2022-emowoz}
Shutong Feng, Nurul Lubis, Christian Geishauser, Hsien-chin Lin, Michael Heck, Carel van Niekerk, and Milica Gasic. 2022.
\newblock \href {https://aclanthology.org/2022.lrec-1.436} {{E}mo{WOZ}: A large-scale corpus and labelling scheme for emotion recognition in task-oriented dialogue systems}.
\newblock In \emph{Proceedings of the Thirteenth Language Resources and Evaluation Conference}, pages 4096--4113, Marseille, France. European Language Resources Association.

\bibitem[{Gehrmann et~al.(2019)Gehrmann, Strobelt, and Rush}]{25-gehrmann2019}
Sebastian Gehrmann, Hendrik Strobelt, and Alexander Rush. 2019.
\newblock \href {https://doi.org/10.18653/v1/P19-3019} {{GLTR}: Statistical detection and visualization of generated text}.
\newblock In \emph{Proceedings of the 57th Annual Meeting of the Association for Computational Linguistics: System Demonstrations}, pages 111--116, Florence, Italy. Association for Computational Linguistics.

\bibitem[{Guo et~al.(2023)Guo, Zhang, Wang, Jiang, Nie, Ding, Yue, and Wu}]{31-guo2023}
Biyang Guo, Xin Zhang, Ziyuan Wang, Minqi Jiang, Jinran Nie, Yuxuan Ding, Jianwei Yue, and Yupeng Wu. 2023.
\newblock \href {https://arxiv.org/abs/2301.07597} {How close is chatgpt to human experts? comparison corpus, evaluation, and detection}.
\newblock \emph{Preprint}, arXiv:2301.07597.

\bibitem[{Hans et~al.(2024)Hans, Schwarzschild, Cherepanova, Kazemi, Saha, Goldblum, Geiping, and Goldstein}]{binoculars}
Abhimanyu Hans, Avi Schwarzschild, Valeriia Cherepanova, Hamid Kazemi, Aniruddha Saha, Micah Goldblum, Jonas Geiping, and Tom Goldstein. 2024.
\newblock Spotting llms with binoculars: zero-shot detection of machine-generated text.
\newblock In \emph{Proceedings of the 41st International Conference on Machine Learning}, pages 17519--17537.

\bibitem[{Hays et~al.(2023)Hays, Schutzman, Raghavan, Walk, and Zimmer}]{65-Hays_2023}
Chris Hays, Zachary Schutzman, Manish Raghavan, Erin Walk, and Philipp Zimmer. 2023.
\newblock \href {https://doi.org/10.1145/3543507.3583214} {Simplistic collection and labeling practices limit the utility of benchmark datasets for twitter bot detection}.
\newblock In \emph{Proceedings of the ACM Web Conference 2023}, WWW ’23. ACM.

\bibitem[{Kirchenbauer et~al.(2023)Kirchenbauer, Geiping, Wen, Katz, Miers, and Goldstein}]{40-kirchenbauer2024}
John Kirchenbauer, Jonas Geiping, Yuxin Wen, Jonathan Katz, Ian Miers, and Tom Goldstein. 2023.
\newblock A watermark for large language models.
\newblock In \emph{International Conference on Machine Learning}, pages 17061--17084. PMLR.

\bibitem[{Koike et~al.(2024)Koike, Kaneko, and Okazaki}]{64-Koike-2024}
Ryuto Koike, Masahiro Kaneko, and Naoaki Okazaki. 2024.
\newblock \href {https://doi.org/10.1609/aaai.v38i19.30120} {Outfox: Llm-generated essay detection through in-context learning with adversarially generated examples}.
\newblock \emph{Proceedings of the AAAI Conference on Artificial Intelligence}, 38(19):21258--21266.

\bibitem[{Kojima et~al.(2022)Kojima, Gu, Reid, Matsuo, and Iwasawa}]{59-kojima2023}
Takeshi Kojima, Shixiang~Shane Gu, Machel Reid, Yutaka Matsuo, and Yusuke Iwasawa. 2022.
\newblock Large language models are zero-shot reasoners.
\newblock \emph{Advances in neural information processing systems}, 35:22199--22213.

\bibitem[{Krishna et~al.(2024)Krishna, Song, Karpinska, Wieting, and Iyyer}]{49-krishna2023}
Kalpesh Krishna, Yixiao Song, Marzena Karpinska, John Wieting, and Mohit Iyyer. 2024.
\newblock Paraphrasing evades detectors of ai-generated text, but retrieval is an effective defense.
\newblock \emph{Advances in Neural Information Processing Systems}, 36.

\bibitem[{Labruna et~al.(2023)Labruna, Brenna, Zaninello, and Magnini}]{51-labruna2023}
Tiziano Labruna, Sofia Brenna, Andrea Zaninello, and Bernardo Magnini. 2023.
\newblock Unraveling chatgpt: A critical analysis of ai-generated goal-oriented dialogues and annotations.
\newblock In \emph{International Conference of the Italian Association for Artificial Intelligence}, pages 151--171. Springer.

\bibitem[{Lavergne et~al.(2008)Lavergne, Urvoy, and Yvon}]{24-Lavergne-2008}
Thomas Lavergne, Tanguy Urvoy, and Fran\c{c}ois Yvon. 2008.
\newblock Detecting fake content with relative entropy scoring.
\newblock In \emph{Proceedings of the 2008 International Conference on Uncovering Plagiarism, Authorship and Social Software Misuse - Volume 377}, PAN'08, page 27–31, Aachen, DEU. CEUR-WS.org.

\bibitem[{Lewis et~al.(2020)Lewis, Liu, Goyal, Ghazvininejad, Mohamed, Levy, Stoyanov, and Zettlemoyer}]{44-lewis-etal-2020}
Mike Lewis, Yinhan Liu, Naman Goyal, Marjan Ghazvininejad, Abdelrahman Mohamed, Omer Levy, Veselin Stoyanov, and Luke Zettlemoyer. 2020.
\newblock \href {https://doi.org/10.18653/v1/2020.acl-main.703} {{BART}: Denoising sequence-to-sequence pre-training for natural language generation, translation, and comprehension}.
\newblock In \emph{Proceedings of the 58th Annual Meeting of the Association for Computational Linguistics}, pages 7871--7880, Online. Association for Computational Linguistics.

\bibitem[{Li et~al.(2024)Li, He, Bai, and Wen}]{66-Li_2024}
Yupeng Li, Haorui He, Jin Bai, and Dacheng Wen. 2024.
\newblock \href {https://doi.org/10.1145/3589334.3645385} {Mcfend: A multi-source benchmark dataset for chinese fake news detection}.
\newblock In \emph{Proceedings of the ACM Web Conference 2024}, volume~9 of \emph{WWW ’24}, page 4018–4027. ACM.

\bibitem[{Lu et~al.(2024)Lu, Liu, Yu, Li, and King}]{39-lu2024}
Yijian Lu, Aiwei Liu, Dianzhi Yu, Jingjing Li, and Irwin King. 2024.
\newblock \href {https://doi.org/10.18653/v1/2024.acl-long.630} {An entropy-based text watermarking detection method}.
\newblock In \emph{Proceedings of the 62nd Annual Meeting of the Association for Computational Linguistics (Volume 1: Long Papers)}, pages 11724--11735, Bangkok, Thailand. Association for Computational Linguistics.

\bibitem[{Mao et~al.()Mao, Vondrick, Wang, and Yang}]{50-mao2024}
Chengzhi Mao, Carl Vondrick, Hao Wang, and Junfeng Yang.
\newblock Raidar: generative ai detection via rewriting.
\newblock In \emph{The Twelfth International Conference on Learning Representations}.

\bibitem[{Mindner et~al.(2023)Mindner, Schlippe, and Schaaff}]{36-Mindner-2023}
Lorenz Mindner, Tim Schlippe, and Kristina Schaaff. 2023.
\newblock Classification of human- and ai-generated texts: Investigating features for chatgpt.
\newblock In \emph{Artificial Intelligence in Education Technologies: New Development and Innovative Practices}, pages 152--170, Singapore. Springer Nature Singapore.

\bibitem[{Mitchell et~al.(2023)Mitchell, Lee, Khazatsky, Manning, and Finn}]{26-Mitchell-2023}
Eric Mitchell, Yoonho Lee, Alexander Khazatsky, Christopher~D Manning, and Chelsea Finn. 2023.
\newblock \href {https://proceedings.mlr.press/v202/mitchell23a.html} {{D}etect{GPT}: Zero-shot machine-generated text detection using probability curvature}.
\newblock In \emph{Proceedings of the 40th International Conference on Machine Learning}, volume 202 of \emph{Proceedings of Machine Learning Research}, pages 24950--24962. PMLR.

\bibitem[{Okur et~al.(2022)Okur, Sahay, and Nachman}]{47-okur2022}
Eda Okur, Saurav Sahay, and Lama Nachman. 2022.
\newblock Data augmentation with paraphrase generation and entity extraction for multimodal dialogue system.
\newblock In \emph{Proceedings of the Thirteenth Language Resources and Evaluation Conference}, pages 4114--4125.

\bibitem[{OpenAI(2023)}]{openai2023gpt35}
OpenAI. 2023.
\newblock Gpt-3.5: Openai's generative pre-trained transformer 3.5.
\newblock \url{https://platform.openai.com}.

\bibitem[{{OWASP Foundation}(2023)}]{61-owasp-2023}
{OWASP Foundation}. 2023.
\newblock Owasp top 10 for large language model applications.
\newblock \href{https://owasp.org/www-project-top-10-for-large-language-model-applications/}{https://owasp.org/www-project-top-10-for-large-language-model-applications/}(Accessed: 2024-09-02).

\bibitem[{Rastogi et~al.(2020)Rastogi, Zang, Sunkara, Gupta, and Khaitan}]{3-rastogi-2020}
Abhinav Rastogi, Xiaoxue Zang, Srinivas Sunkara, Raghav Gupta, and Pranav Khaitan. 2020.
\newblock Towards scalable multi-domain conversational agents: The schema-guided dialogue dataset.
\newblock In \emph{Proceedings of the AAAI conference on artificial intelligence}, volume~34, pages 8689--8696.

\bibitem[{Reimers and Gurevych(2019)}]{68-reimers-2019}
Nils Reimers and Iryna Gurevych. 2019.
\newblock \href {https://doi.org/10.18653/v1/D19-1410} {Sentence-{BERT}: Sentence embeddings using {S}iamese {BERT}-networks}.
\newblock In \emph{Proceedings of the 2019 Conference on Empirical Methods in Natural Language Processing and the 9th International Joint Conference on Natural Language Processing (EMNLP-IJCNLP)}, pages 3982--3992, Hong Kong, China. Association for Computational Linguistics.

\bibitem[{Sennrich et~al.(2016)Sennrich, Haddow, and Birch}]{43-sennrich-etal-2016}
Rico Sennrich, Barry Haddow, and Alexandra Birch. 2016.
\newblock \href {https://doi.org/10.18653/v1/P16-1009} {Improving neural machine translation models with monolingual data}.
\newblock In \emph{Proceedings of the 54th Annual Meeting of the Association for Computational Linguistics (Volume 1: Long Papers)}, pages 86--96, Berlin, Germany. Association for Computational Linguistics.

\bibitem[{Taud and Mas(2018)}]{MLP-Taud2018}
H.~Taud and J.F. Mas. 2018.
\newblock \href {https://doi.org/10.1007/978-3-319-60801-3_27} {\emph{Multilayer Perceptron (MLP)}}, pages 451--455.
\newblock Springer International Publishing, Cham.

\bibitem[{Team et~al.(2024)Team, Georgiev, Lei, Burnell, Bai, Gulati, Tanzer, Vincent, Pan, Wang et~al.}]{73-2024gemini}
Gemini Team, Petko Georgiev, Ving~Ian Lei, Ryan Burnell, Libin Bai, Anmol Gulati, Garrett Tanzer, Damien Vincent, Zhufeng Pan, Shibo Wang, and 1 others. 2024.
\newblock Gemini 1.5: Unlocking multimodal understanding across millions of tokens of context.
\newblock \emph{arXiv preprint arXiv:2403.05530}.

\bibitem[{Touvron et~al.(2023)Touvron, Lavril, Izacard et~al.}]{touvron2023llama}
Hugo Touvron, Thibaut Lavril, Gautier Izacard, and 1 others. 2023.
\newblock Llama: Open and efficient foundation language models.
\newblock \url{https://ai.facebook.com/blog/large-language-model-llama-meta-ai/}.

\bibitem[{Wahle et~al.(2022)Wahle, Ruas, Kirstein, and Gipp}]{Wahle-2022}
Jan~Philip Wahle, Terry Ruas, Frederic Kirstein, and Bela Gipp. 2022.
\newblock \href {https://doi.org/10.18653/v1/2022.emnlp-main.62} {How large language models are transforming machine-paraphrase plagiarism}.
\newblock In \emph{Proceedings of the 2022 Conference on Empirical Methods in Natural Language Processing}, pages 952--963, Abu Dhabi, United Arab Emirates. Association for Computational Linguistics.

\bibitem[{Wang et~al.(2023)Wang, Li, Ren, Jiang, Zhang, and Qiu}]{29-Wang-2023}
Pengyu Wang, Linyang Li, Ke~Ren, Botian Jiang, Dong Zhang, and Xipeng Qiu. 2023.
\newblock \href {https://aclanthology.org/2023.emnlp-main.73} {{S}eq{XGPT}: Sentence-level {AI}-generated text detection}.
\newblock In \emph{Proceedings of the 2023 Conference on Empirical Methods in Natural Language Processing}, pages 1144--1156, Singapore. Association for Computational Linguistics.

\bibitem[{Zhan et~al.(2024)Zhan, Liang, Ying, and Kang}]{Zhan-2024}
Qiusi Zhan, Zhixiang Liang, Zifan Ying, and Daniel Kang. 2024.
\newblock \href {https://aclanthology.org/2024.findings-acl.624} {{I}njec{A}gent: Benchmarking indirect prompt injections in tool-integrated large language model agents}.
\newblock In \emph{Findings of the Association for Computational Linguistics ACL 2024}, pages 10471--10506, Bangkok, Thailand and virtual meeting. Association for Computational Linguistics.

\bibitem[{Zheng et~al.(2023)Zheng, Chiang, Sheng, Li, Zhuang, Wu, Zhuang, Li, Lin, Xing et~al.}]{6-zheng-2024}
Lianmin Zheng, Wei-Lin Chiang, Ying Sheng, Tianle Li, Siyuan Zhuang, Zhanghao Wu, Yonghao Zhuang, Zhuohan Li, Zi~Lin, Eric~P Xing, and 1 others. 2023.
\newblock Lmsys-chat-1m: A large-scale real-world llm conversation dataset.
\newblock \emph{CoRR}.

\bibitem[{Zhong et~al.(2022)Zhong, Liu, Yin, Mao, Jiao, Liu, Zhu, Ji, and Han}]{72-zhong2022}
Ming Zhong, Yang Liu, Da~Yin, Yuning Mao, Yizhu Jiao, Pengfei Liu, Chenguang Zhu, Heng Ji, and Jiawei Han. 2022.
\newblock Towards a unified multi-dimensional evaluator for text generation.
\newblock In \emph{2022 Conference on Empirical Methods in Natural Language Processing, EMNLP 2022}.

\bibitem[{Zhu et~al.(2023)Zhu, Geishauser, Lin, van Niekerk, Peng, Zhang, Feng, Heck, Lubis, Wan et~al.}]{5-zhu-2023}
Qi~Zhu, Christian Geishauser, Hsien-Chin Lin, Carel van Niekerk, Baolin Peng, Zheng Zhang, Shutong Feng, Michael Heck, Nurul Lubis, Dazhen Wan, and 1 others. 2023.
\newblock Convlab-3: A flexible dialogue system toolkit based on a unified data format.
\newblock In \emph{Proceedings of the 2023 Conference on Empirical Methods in Natural Language Processing: System Demonstrations}, pages 106--123.

\bibitem[{Zhu et~al.(2020)Zhu, Huang, Zhang, Zhu, and Huang}]{10-zhu-etal-2020}
Qi~Zhu, Kaili Huang, Zheng Zhang, Xiaoyan Zhu, and Minlie Huang. 2020.
\newblock \href {https://doi.org/10.1162/tacl_a_00314} {{C}ross{WOZ}: A large-scale {C}hinese cross-domain task-oriented dialogue dataset}.
\newblock \emph{Transactions of the Association for Computational Linguistics}, 8:281--295.

\end{thebibliography}
\newpage
\appendix

\section{Appendix}
\subsection{Prompt Example for Data Augmentation Framework}\label{prompt example structure}

In this section, we provide examples of prompt settings based on a sample full-human dialogue for all components of the data augmentation framework, including eight tables that illustrate sample prompts for each of our proposed data augmentation frameworks.

Table \ref{tab:full human example dialogue} shows the goal and original Full-Human dialogue of our example. 
Table \ref{tab:Llama 70B missing sentence completion prompt example part 1} and Table \ref{tab:Llama 70B missing sentence completion prompt example part 2} show the Llama 70B structured prompt example for Missing Sentence Completion with a consist of Task, Slot and Domain Knowledge, Chain of thought and Chat history with missing sentences.
Table \ref{tab:GPT3.5 missing sentence completion prompt example} shows the example prompt for GPT-3.5 used in the Missing Sentence Completion framework.
Table \ref{tab:Next Response Generation prompt example} provides an example prompt for Next Response Generation, which follows the same structure for both GPT-3.5 and Llama 70B.
Table \ref{tab: par. prompt example} shows the two staged prompt for Par. framework.
Table \ref{tab: G2D prompt example} and Table \ref{tab: G2D prompt example part 2} show the example prompt for G2D framework.
Table \ref{tab: E2E Convo. GPT3.5 user prompt example} and Table \ref{tab: E2E Convo. GPT3.5 system prompt example} show an example prompt for GPT-3.5 acting as the user or system in the E2E Convo. framework.
Table \ref{tab: E2E Convo. Llama 70B user prompt example}, Table \ref{tab: E2E Convo. Llama 70B user prompt example part 2} and Table \ref{tab: E2E Convo. Llama 70B system prompt example} show an example prompt for Llama 70B acting as the user or system in the E2E Convo. framework.

\subsection{Multiclass Classification}\label{sec:multiclass classification}
Table~\ref{tab:multiclass-performance} presents the macro F1 scores of detection models on multiclass classification task, which classifies dialogues as generated by human or individual LLMs. Using the new datasets, the labels include ``Human", ``GPT", and ``Llama". Zero-shot detection methods such as Binoculars~\cite{binoculars} targets binary classification only, and is not applicable to the multiclass detection task.

By collapsing all synthetic dialogue sources into a single ``AI" category, binary classification performance (reported in Table~\ref{tab:performance}) consistently outperformed the multiclass classification performance across the same models. For instance, XGBoost, LogR, and SVM all showed higher F1 scores in binary classification than multiclass classification, where distinguishing between multiple LLMs proved more difficult. 

\begin{table*}
    \centering
    \small
    \caption{Macro-F1 of multiclass detection models. The highest score of each detection task is in bold.}
    \resizebox{\textwidth}{!}{%
    \begin{tabular}{llcccccccc}
        \hline
         & & \multicolumn{2}{c}{\textbf{Statistical}} & \multicolumn{1}{c}{\textbf{PLM}} & \multicolumn{5}{c}{\textbf{Feature}} \\
        \hline
            \textbf{Dataset} & \textbf{Detection} & \textbf{Entropy} & \textbf{Binoculars} & \textbf{RoBERTa} & \textbf{MLP} & \textbf{XGboost} & \textbf{LogR} & \textbf{SVM} & \textbf{RF}\\
            \hline
            Par. & Multi-class & 0.5394 & - & 0.9003 & \textbf{0.9483} & 0.8862 & 0.9245 & 0.9352 & 0.9112\\
            \hline
            G2D  & Multi-class & 0.5985 & - & 0.9710 & \textbf{0.9913} & 0.9564 & 0.9835 & 0.9809 & 0.9754\\
            \hline
            E2E Convo. & Multi-class & 0.6019 & - & 0.9708 & \textbf{0.9851} & 0.9561 & 0.9755 & 0.9756 & 0.9634\\
            \hline
            Next Response Generation & Multi-class & 0.5096 & - & 0.8962 & \textbf{0.9161} & 0.8404 & 0.8990 & 0.8916 & 0.8591\\
        \hline
    \end{tabular}
    }
    \label{tab:multiclass-performance}
\end{table*}


\subsection{Full-Chatbot Quality Assurance}\label{full chatbot quality assurance}
For Full-Chatbot dialogues, we additionally measure the degree of match between the goal and the generated dialogue using automated and human survey-based methods. This aims to ensure our generated dialogue does not suffer from the mismatch issues in the original Full-Human dataset as identified in Section~\ref{Fine-tuned MultiWOZ2.1 dataset}. 

\subsubsection{Automated Quality Assurance}\label{automated quality assurance}
This automated goal-dialogue match assurance combined the use of the pre-trained dialogue state tracking model (DST) ~\cite{5-zhu-2023} and the pre-trained sentence similarity model ~\cite{68-reimers-2019}. We input the dialogue and goal separately into the DST model and compare their domain and slot values results using the sentence similarity model, deriving a goal dialogue match score. 

\begin{figure}[h]
  \small
  \centering
  \includegraphics[scale = 0.5]{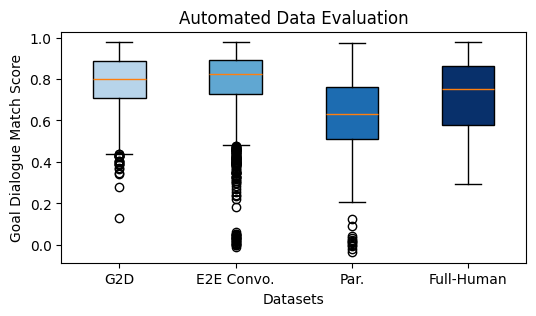}
  \caption{Comparison of goal-dialogue match scores from \underline{Automated Quality Assurance} for dialogues produced by G2D, E2E Convo., Par., and Original Full-Human.}
  \label{fig: data evaluation}
\end{figure}

Figure \ref{fig: data evaluation} shows a boxplot to compare the goal dialogue match score. The result indicates that dialogues generated by G2D and E2E Convo. data augmentation methods have a high average goal dialogue match score of approximately 0.8, while the dialogues generated by the Par. framework show a slightly lower score. This may be due to the framework not providing a dialogue goal in the prompt. To further the findings, we conducted a human survey and asked participants to rank the synthetic dialogues according to the degree of match. Details of the experiments can be found in Appendix \ref{manual quality assurance}. In general, our generated synthetic dialogues are of high quality, as the goal-dialogue match score is similar to that of the Full-Human dialogue dataset, averaging 0.8. 

\subsubsection{Manual Quality Assurance}\label{manual quality assurance}
\begin{figure}[h]
  \centering
  \includegraphics[scale = 0.5]{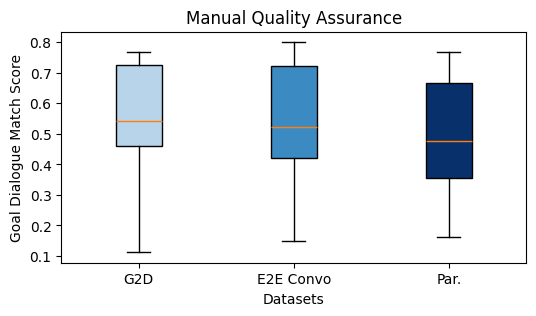}
  \caption{The boxplot comparison using goal-dialogue match scores from  \underline{Manual Quality Assurance }for dialogues generated from G2D, E2E Convo. and, Par..}
  \label{fig: manual_ranking}
\end{figure}

\begin{figure*}[h]
  \centering
  \includegraphics[scale = 0.7]{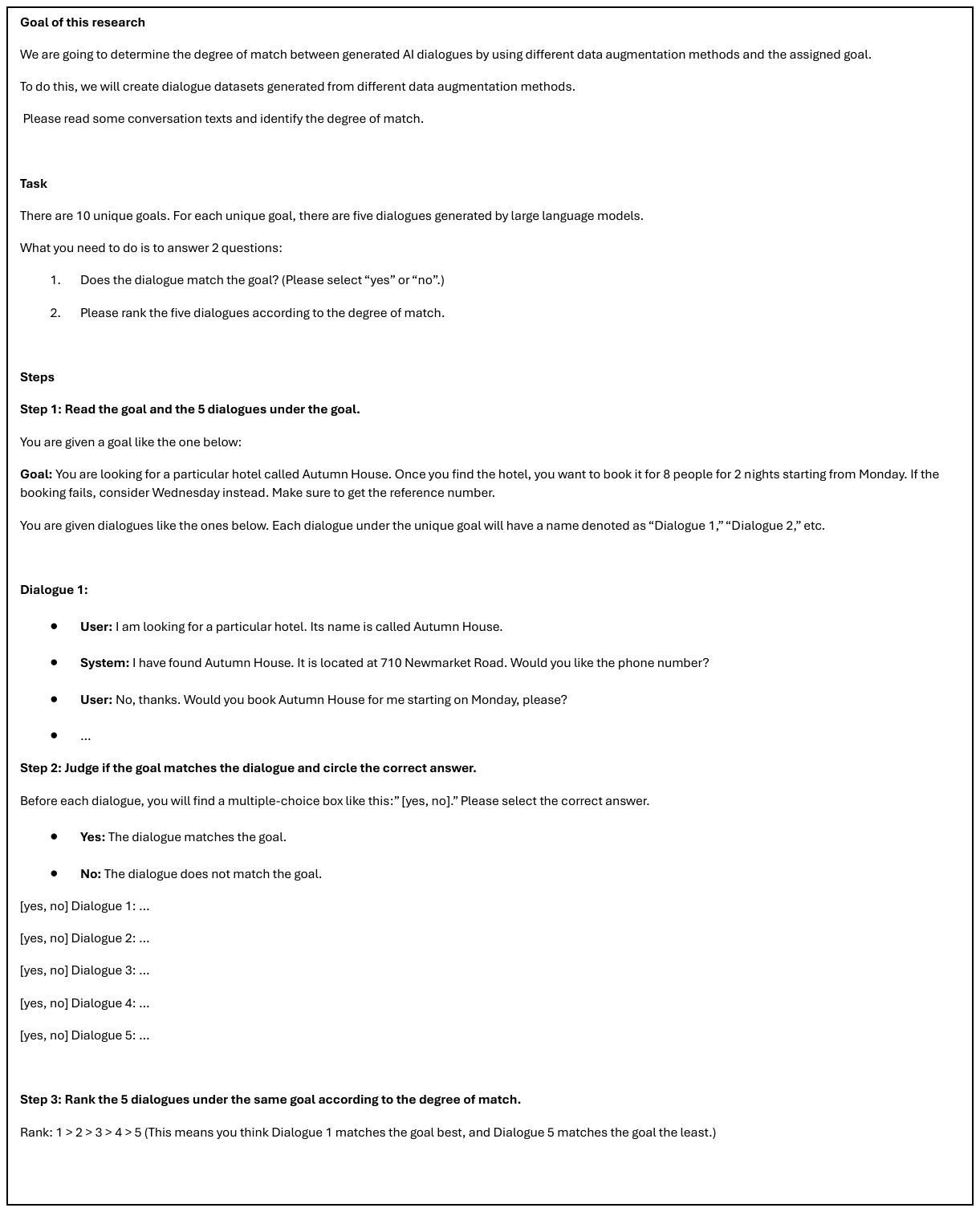}
  \caption{Human survey instruction template.}
  \label{fig: human survey}
\end{figure*}

In the process of Manual Quality Assurance we employ 15 volunteers to complete a survey\footnote{The participants with different highest education levels at university level, which include 5 bachelor students, 8 master students and 2 phd students}. The human survey template is shown in Figure \ref{fig: human survey}, with three main parts: the goal of the research, the task, and the steps.
The survey includes 10 unique goals and 50 synthetic dialogues generated from different data augmentation frameworks. For each unique goal there are 5 dialogues generated using varies data augmentation methods. We inquiry volunteers to complete two tasks: (i) Decide if the given goal match the dialogue by answering `Yes' or `No'. (ii) Ranking 5 dialogues under the same goal according to the degrees of match. During the processing of survey the participants do not know which dialogue is generated from which data augmentation framework, and were ask to filling a google form to complete the survey. 

To define the metric, we calculate the match rate for each dialogue written as: $$\frac{C_{yes}}{N_p}$$ where $C_{yes}$ denotes the number of `yes' responses for the first task, $N_p$ denotes number of participates. Then, we calculate the weighted ranking score for each types of data augmentation framework using the formula: $$\frac{\Sigma_{i=1}^{N_p} R_i}{5 \cdot N_p}$$ where $R$ represents the ranking score for each dialogue. If participants believe that a dialogue best matches the goal, it is recorded as 5, which is the highest score a single dialogue can earn. We multiply the two results calculated above to get a final score to represent the degree of goal dialogue match. 

Figure \ref{fig: manual_ranking} illustrates the goal dialogue match scores for synthetic dialogues generated from our data augmentation frameworks. It shows that Paraphrase dialogues are more likely to be treated as synthetic dialogues with the lowest degree of match between the goal and the generated dialogue, compared to E2E Convo. or G2D dialogues. This supports our conclusion in Appendix~\ref{automated quality assurance} that Paraphrase dialogues have relatively lower goal dialogue match scores.

\subsection{Prompt Exchange Experiment} \label{Prompt Exchange Experiment}
We conducted a prompt exchange experiment to compare the generation capabilities of GPT-3.5 and Llama 70B (Section \ref{Data Collection}). In our experiment, we generated synthetic dialogues using the augmentation frameworks defined in Section \ref{DataAugmentation}, by exchanging prompts between GPT-3.5 and Llama 70B for 20 randomly selected dialogues. For qualitative analysis, we used the end-to-end conversation data augmentation framework as an example. We use Llama 70B user prompt for GPT-3.5 user and conduct E2E Convo. with GPT-3.5 system with true GPT-3.5 system prompt and GPT-3.5 user with Llama 70B user prompt. The generation results show that when a prompt designed for Llama 70B is used with GPT-3.5, the conversation tends to exhibit more formatting issues, such as incorrect information masking and unexpected conversation stops. In contrast, using a GPT-3.5 prompt for Llama 70B leads to fewer formatting issues, but the model becomes more information-hungry and rigidly sticks to the goal, as the system version goal is used in the GPT-3.5 prompt. 

For qualitative analysis, we take the goal from Table \ref{tab:full human example dialogue} as an example. The goal of the user is to book a hotel for 6 people for 3 nights. When using the GPT-3.5 prompt structure on Llama 70B, the output response from the synthetic system will provide information about the "6 people for 3 nights" before the user provides that information. Sample questions include, "I've found a [hotel name] with availability for 6 people for 3 nights; would you like me to confirm the booking?" Since Llama 70B was provided with a system version goal that was modified from the original user goal, it leads to Llama 70B rigidly sticking to the goal when using a GPT-3.5 prompt.

\subsection{Qualitative Analysis for Offline Dialogue Detection}\label{Offline Dialogue Qualitative Analysis}

\begin{figure}[h]
  \centering
  \includegraphics[width=\linewidth]{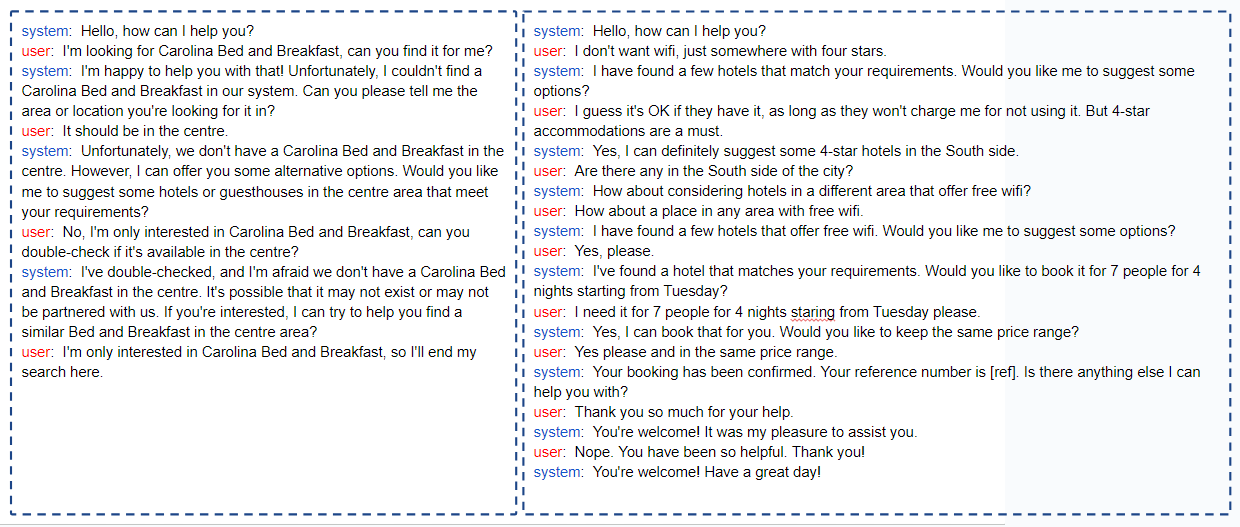}
  \caption{Left: Dialogue with fewer than 5 user utterances, which is easier to misclassification. Right: Dialogue with more than 5 user utterances, typically classified correctly.}
  \label{fig: QA_example}
\end{figure}

In Section \ref{Full Dialogue Detection}, we conduct experiments on synthetic user detection using different models. In this Appendix, we present a sample case of a typical misclassification scenario. From Figure \ref{fig: QA_example}, the left-hand side dialogue is a sample with fewer than 5 user utterances, while the right-hand side dialogue is a sample with more than 5 user utterances. After comparing the misclassified dialogues, we found that errors mostly occurred when the generated dialogues included fewer utterances. These dialogues were sometimes misclassified due to the limited information provided. On the other hand, dialogues with more than five user utterances were seldom misclassified. This suggests that having more chat history provides a better learning base for synthetic user dialogue detection, which also indicates the potential limitation of early detection with limited chat history.

\subsection{Qualitative Analysis for Online Dialogue Detection} \label{qualitative online dialogue detection}
In this section we provide an example of misclassification when there is only one user utterance available. 
When LLM-users start a conversation, their opening sentence can be similar to human users, making the detection harder. For example, users attempting to find a particular hotel in the E2E Convo. dataset is commonly misclassified by RoBERTa when only 1 utterance is available, but correctly classified afterwards. Here is a llama-user classified as human:  
"user: Hello, I'm looking for Bridge Guest House. Can you help me find it?"
A possible reason is that such utterances contain minimal information, and is very similar to how humans start similar dialogues.
We can identify opening sentences made by human users that exhibit small differences to the misclassified example: 
"user: Hello, I am looking for a hotel called the worth house. Can you help me find it?"
This reinforces the limitation of early synthetic dialogue detection in an online environment, as introduced in Section \ref{Offline Dialogue Qualitative Analysis}.

\subsection{Online Dialogue Detection Enhancement}\label{appendix:online_dia}

\begin{table}[h]
    \centering
    \small
    \caption{Model Performance on Par., G2D, E2E Convo. with stacking or expanded dataset strategies. Performance on exactly the number of user utterances specified, and dialogues with fewer user utterances are excluded.}
    \resizebox{\columnwidth}{!}{%
    \begin{tabular}{llccc}
    \hline
         & & \multicolumn{1}{c}{\textbf{Stack}} & \multicolumn{2}{c}{\textbf{Expanded Dataset}} \\
        \hline
            \textbf{Dataset} & \textbf{\#Utterance} & Stack & \textbf{$ExpPro_{MLP}$} & \textbf{$ExpPro_{RoBERTa}$}\\
            \hline
            Par. & 1 & 0.6908 & \textbf{0.7479} & 0.6959 \\
            \textbf{} & 2 & 0.8170 & 0.8286 & \textbf{0.8612} \\
            \textbf{} & 3 & 0.8855 & 0.8085 & \textbf{0.8919}\\
            \textbf{} & 4 & \textbf{0.9231} & 0.9130 & 0.8927 \\
            \hline
            G2D & 1 & 0.8865 & 0.7577 & \textbf{ 0.9028}\\
            \textbf{} & 2 & 0.8997 & 0.6703 & \textbf{0.9238} \\
            \textbf{} & 3 & \textbf{0.9382} & 0.9197 & 0.9263 \\
            \textbf{} & 4& \textbf{0.9668} & 0.9644 & 0.9533 \\
            \hline
            E2E Convo. & 1 & 0.9373 & 0.8028 & \textbf{0.9792} \\
            \textbf{} & 2 & 0.9670 & 0.8542 & \textbf{0.9901} \\
            \textbf{} & 3 & 0.9874 & 0.8667 & \textbf{0.9944} \\
            \textbf{} & 4 & 0.9923 & 0.8373 & \textbf{0.9996} \\
        \hline
    \end{tabular}
    }
    \label{tab:progressing_performance_enhancement}
\end{table}

We found two potential frameworks that are previously used in a similar situation to enhance the detection performance on incomplete dialogues, which are stacking, and dataset expansion.

\subsubsection{Stacking}
Feature based models $M_F$ and $RoBERTa$ model receive different input and features for detection, and it is hypothesized that more comprehensive detection can be made when the two approaches are combined via a stacking classifier $M_S$. The following procedure aims to train a system with enhanced performance on detection given $n$ user utterances: 
\begin{itemize}
    \item Given training dataset $D_T$ and labels $y_T$
    \item Train base classifiers $M_F$ and $RoBERTa$ on $D_T$ and $y_T$
    \item Randomly sample $D_T' \subseteq D_T$ and extract the first $n$ user utterances from each dialogue to form a stacking dataset $D_S$ with labels $y_S$
    \item $M_F$ and $RoBERTa$ make inferences on $D_S$, constructing $P_S$ which consists of $RoBERTa$ predictions (average of per-word probabilities in the utterances) and $M_F$ logits
    \item Train $M_S$ on $P_S$ and $y_S$\\
\end{itemize}
Then to detect a dialogue $d_i$ of $n$ user utterances:
\begin{itemize}
    \item Base classifiers $M_F$ and $RoBERTa$ make inferences on $d_i$, giving $p_i$ consisting of $RoBERTa$ predictions and $M_F$ logits
    \item $M_S$ makes inferences on $p_i$ to produce final prediction
\end{itemize}

\subsubsection{Dataset Expansion Approach}
We expand the training dataset for progressive dialogue detection via $ExpPro$, specifically, for each complete dialogue $d_i$ with $N_i$ turns, crop the dialogue at first $k \in \mathbb{Z}^+ < N_i$ turns and include them into the training dataset.

The method is applied on the feature-based MLP and pre-trained RoBERTa, which are the top-performing classifiers.

\subsubsection{Experiment and Results}

We conducted our experiments using robust performance models MLP and RoBERTa, as indicated in Section \ref{Full Dialogue Detection}, based on the two defined methods described above. We performed comparison experiments on E2E Convo. dataset with progressive dialogue, using a varying number of user utterances as input, having label of either `human' or `AI'.

In our experiments, we utilize user utterances ranging from 1 to 4, as more than half of the dialogues do not contain more than 4 user utterances, which would not represent the majority of cases. In Table \ref{tab:progressing_performance_enhancement}, the F1 scores of each experiment are recorded. We found that RoBERTa, trained on expanded datasets that include all possible cases of chat history, demonstrated the best performance, showing a significant improvement compared to the original results trained on full dialogues and tested on progressive dialogues. The results indicate that common performance improvement methods are effective for progressive dialogue detection.

\subsection{Out-of-Distribution Detection}\label{appendix:external-data}
\begin{table}[h]
    \centering
    \small
    \caption{RoBERTa's Macro-F1 performance on SGD and CrossWOZ datasets augmented using G2D framework, when trained on different datasets based on MultiWOZ dataset.}
    \begin{tabular}{lcc}
    \hline
    \textbf{Train Dataset} & \textbf{SGD} & \textbf{CrossWOZ} \\
    \hline
    P & 0.3333 & 0.3333 \\
    E & 0.5016 & 0.5238 \\
    G & 0.4286 & 0.3960 \\
    \hline
    Single Average & 0.4212 & 0.4177 \\
    \hline
    P + E & 0.6328 & 0.5572 \\
    P + G & 0.7971 & 0.4956 \\
    E + G & 0.3984 & 0.5696 \\
    \hline
    Combination Average & 0.6094 & 0.5408 \\
    \hline
    \end{tabular}
    \label{tab:ood-detection}
\end{table}

To further demonstrate that integrating multiple augmentation frameworks enhances generalizability, experiments are conducted using subsets of two external datasets featuring customer service line dialogues with human user: SGD~\cite{3-rastogi-2020} and CrossWOZ~\cite{10-zhu-etal-2020}. In addition, a different augmentor, Gemini 1.5~\cite{73-2024gemini}, is employed to perform the G2D augmentation for synthetic user dialogue generation. A RoBERTa model is trained on the datasets based on MultiWoz and tested on the additional datasets. During the experiments, we control all training set sizes (1232 dialogues) when training RoBERTa on a single or a combination of frameworks, ensure consistent testing datasets across training scenarios, and ensure consistent datasets in terms of masked information across positive and negative samples (e.g. [hotel name]), which is a more challenging scenario. 

The RoBERTa Macro-F1 results are presented in Table~\ref{tab:ood-detection}. Despite the performance decline caused by differences in the external testing dataset compared to our training dataset, the use of a different LLM, and the fact that the CrossWOZ dataset used is translated into English and typically features short dialogues (fewer than 5 turns), the results across both additional and our own datasets consistently demonstrate that integrating multiple frameworks enhances generalisability.

\subsection{Data Augmentation Framework Demonstration}
\begin{figure*}[h]
  \centering
  \includegraphics[width=\textwidth]{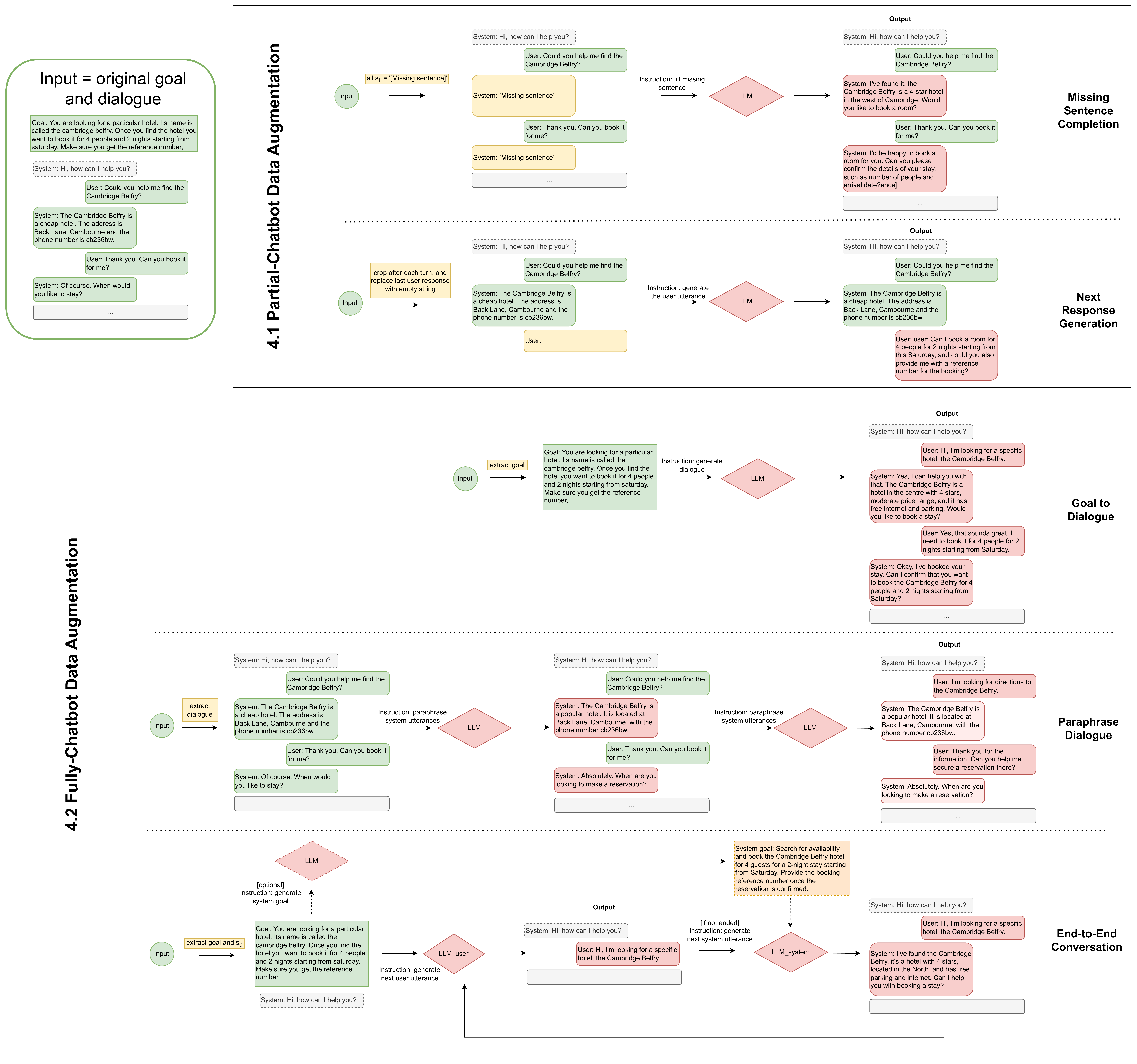}
  \caption{Demonstration of the data augmentation frameworks. The initial system response is set to a standard starting line common to all dialogues. The instructions provided to LLMs as depicted in the figure are incomplete.}
  \label{fig:framework-demo}
\end{figure*}

In this section, a demonstration of our proposed data augmentation frameworks is illustrated in Figure \ref{fig:framework-demo}.  The demonstration takes a dialogue as an example and shows the exact process of augmentation for the dialogue system, including all relevant queries in our framework.

\begin{table*}[ht]
    \centering
    \small
    \caption{Example for Full-Human dialogue with goal.}
    \resizebox{\textwidth}{!}{%
    \begin{tabular}{p{2cm}p{15cm}}
    \hline
        \textbf{Components} & \textbf{Prompt}\\
    \hline
    Goal & The user is looking for a place to stay. The hotel should be in the cheap price range and should be in the type of hotel. The hotel should include free parking and should include free wifi. Once the user find the hotel the user want to book it for 6 people and 3 nights starting from tuesday. If the booking fails how about 2 nights. Make sure the uer get the reference number.\\
    Chat History & user: am looking for a place to to stay that has cheap price range it should be in a type of hotel. \textbackslash n\\ 
    &system: Okay, do you have a specific area you want to stay in? \textbackslash n\\
    &user: no, i just need to make sure it's cheap. oh, and i need parking. \textbackslash n\\ 
    &system: I found 1 cheap hotel for you that includes parking. Do you like me to book it? \textbackslash n\\
    \hline
    \end{tabular}
    }
    \label{tab:full human example dialogue}
\end{table*}

\begin{table*}[t]
    \centering
    \small
    \caption{Structured user prompt example for Missing Sentence Completion data augmentation for Llama 70B.}
    \resizebox{\textwidth}{!}{%
    \begin{tabular}{p{2cm}p{15cm}}
    \hline
        \textbf{Components} & \textbf{Prompt}\\

        \hline
        Task & Task: Replace each of the "[missing sentence]" in the dialogue.\\

        Slot and Domain Knowledge & "internet": \{ "description": "whether the hotel has internet", "is\_categorical": true, "possible\_values": $\left[ \text{"free"}, \text{"no"}, \text{"yes"} \right]$ \}\\
         &"parking": \{ "description": "whether the hotel has parking", "is\_categorical": true, "possible\_values": $\left[ \text{"free"}, \text{"no"}, \text{"yes"} \right]$ \},\\
        & "area": \{ "description": "area or place of the hotel", "is\_categorical": true, "possible\_values": $\left[\text{"centre"}, \text{"east"}, \text{"north"}, \text{"south"}, \text{"west"} \right]$ \}\\
        & "stars": \{ "description": "star rating of the hotel", "is\_categorical": true, "possible\_values":  $\left[ \text{"0"}, \text{"1"}, \text{"2"}, \text{"3"}, \text{"4"}, \text{"5"} \right]$ \}\\
        & "price range": \{"description": "price budget of the hotel", "is\_categorical": true, "possible\_values": $\left[\text{"expensive"}, \text{"cheap"}, \text{"moderate"} \right]$ \}\\
        & "type": \{"description": "what is the type of the hotel", "is\_categorical": true, "possible\_values": $\left[\text{"guesthouse"}, \text{"hotel"} \right]$ \}\\
        & "name": \{ "description": "name of the hotel", "is\_categorical": false, "possible\_values": $\left[ \right]$ \}\\
        & "book people": \{ "description": "number of people for the hotel booking", "is\_categorical": false, "possible\_values": $\left[ \right]$ \}\\
        & "book stay": \{ "description": "length of stay at the hotel", "is\_categorical": false, "possible\_values": $\left[ \right]$ \}\\
        & "book day": \{ "description": "day of the hotel booking", "is\_categorical": true, "possible\_values": $\left[\text{"monday"}, \text{"tuesday"}, \text{"wednesday"}, \text{"thursday"}, \text{"friday"}, \text{"saturday"}, \text{"sunday"} \right]$ \}\\
        & "phone": \{ "description": "phone number of the hotel", "is\_categorical": false, "possible\_values": $\left[ \right]$ \}\\
        & "postcode": \{ "description": "postcode of the hotel", "is\_categorical": false, "possible\_values": $\left[ \right]$ \}\\
        & "address": \{ "description": "address of the hotel", "is\_categorical": false, "possible\_values": $\left[ \right]$ \}\\
        & "ref": \{ "description": "reference number of the hotel booking", "is\_categorical": false, "possible\_values": $\left[ \right]$ \}\\
        & "choice": \{ "description": "number of hotels that meet the requirement", "is\_categorical": false, "possible\_values": $\left[ \right]$ \}\\
            Chain of Thought &For each missing sentence, your response should be in format of: \\
&turn\_id\\
&- impact of immediately preceding user sentence\\
&- impact of immediately following user sentence (note that the real system only have knowledge up to the missing sentence)\\
&- impact of overall previous and following context\\
&- one line replacing the missing sentence\\
&Afterall, print the completed entire dialogue.\\
&Here is a demonstration of the task where response is bounded by \\
&==========: \\
        \hline
    \end{tabular}
    }
    \label{tab:Llama 70B missing sentence completion prompt example part 1}
\end{table*}

\begin{table*}[t]
    \centering
    \small
    \caption{Structured user prompt example for Missing Sentence Completion data augmentation for Llama 70B (continuous Table \ref{tab:Llama 70B missing sentence completion prompt example part 1}).}
    \resizebox{\textwidth}{!}{%
    \begin{tabular}{p{2cm}p{15cm}}
    \hline
        \textbf{Components} & \textbf{Prompt}\\

        \hline

Chain of Thought &user: I'm looking for a hotel to stay at in the centre, can you look this up for me?\\
&system: [missing sentence]\\
&user: Not in terms of that, but do they have free parking and have a 3 star rating?\\
&system: [missing sentence]\\
&user: Okay, I'd like to book a room at the Gonville Hotel for 4 nights. There will be 6 people and we will be arriving on Saturday.\\
&system: [missing sentence]\\
&user: Yes, what about 2 nights instead of 4?\\
&system: [missing sentence]\\
&user: No, that looks like everything.  Thanks.  Bye.\\
&system: [missing sentence]\\
&==========\\
&1. \\
&- the user asks the system to look up a hotel in the centre, system should respond to this query\\
&- "not in terms of that" seems to be responding to the system's suggestion which the user does not care about (e.g. pricerange, free wifi that are not required by user anywhere in the dialogue). the user asks if "they" have free parking and 3 star rating, which means the system should provid some hotel suggestions, but NOT parking and star information!\\
&- later in the chat the user mentioned Gonville hotel which the system has likely suggested to them. If suggesting particular hotels, it is likely one of them.\\
&- There are three hotels in the center of town. Do you prefer something moderate or expensive? \\
&2. \\
&- the user asked if the suggested hotels has free parking and is 3 star which the system must respond to\\
&- the user replies "okay" to the system's suggestion and provided details of their booking at Gonville Hotel. The system has likely suggested Gonville Hotel and asked if the user wish to make a booking\\
&- NA\\
&- The Gonville hotel has 3 stars and parking, and the University Arms hotel has 4 stars and parking. They are both expensive. Would you like more details? \\
&3.\\ 
&- booking details at a particular hotel has been provided by the user, the system can attempt to make a booking\\
&- the user says 'yes' to system and then shortend the stay to 2 nights. Which means the system was potentially unable to make booking as initially required and suggested to shorten the days.\\
&- Book day, book stay, book people must be provided to make a booking. In case of failed booking, system can suggest to change length or time of stay, or suggest another hotel that satisfy all requirement, or ask user to relax a previously stated requirement (in this case 3 star, free parking, area)\\
&- I'm sorry, there are no rooms available for that length of stay.  Could you shorten your stay or book a different day possibly?\\
&4. \\
&- the user agreed to shorten the stay\\
&- the user says 'no' to system and claims that they are all set and ended the conversation. The system has likely successfully make a booking and asked if user need anything else.\\
&- NA\\
&- Sure, that worked.  You have booked 2 nights and your reference number is RU89U6V8.  Can I be of further help today?\\
&5.\\
&- the user expressed appreciation and ended the conversation\\
&- no more user sentence.\\
&- user has ended conversation. system should end conversation\\
&- You're welcome. Enjoy your stay!\\
&Completed dialogue: \\
&user: I'm looking for a hotel to stay at in the centre, can you look this up for me?\\
&system: There are three hotels in the center of town. Do you prefer something moderate or expensive?\\ 
&user: Not in terms of that, but do they have free parking and have a 3 star rating?\\
&system: The Gonville hotel has 3 stars and parking, and the University Arms hotel has 4 stars and parking. They are both expensive. Would you like more details? \\
&user: Okay, I'd like to book a room at the Gonville Hotel for 4 nights. There will be 6 people and we will be arriving on Saturday. \\
&system: I'm sorry, there are no rooms available for that length of stay.  Could you shorten your stay or book a different day possibly?\\
&user: Yes, what about 2 nights instead of 4?\\
&system: Sure, that worked.  You have booked 2 nights and your reference number is RU89U6V8.  Can I be of further help today?\\
&user: No, that looks like everything.  Thanks.  Bye.\\
&system: You're welcome. Enjoy your stay!\\
&==========\\

Chat history&Here is the dialogue of your task:\\
&user: am looking for a place to to stay that has cheap price range it should be in a type of hotel. \textbackslash n\\ 
        &System: [missing sentence] \textbackslash n\\
        &user: no, i just need to make sure it's cheap. oh, and i need parking. \textbackslash n\\ 
        &System: [missing sentence] \textbackslash n\\
        \hline

    \end{tabular}
    }
    \label{tab:Llama 70B missing sentence completion prompt example part 2}
\end{table*}

\begin{table*}[t]
    \centering
    \small
    \caption{Structured user prompt example for Missing Sentence Completion data augmentation for GPT3.5.}
    \resizebox{\textwidth}{!}{%
    \begin{tabular}{lp{15cm}}
    \hline
        \textbf{Components} & \textbf{Prompt}\\
        \hline
         Goal& goal: The user is looking for a place to stay. The hotel should be in the cheap price range and should be in the type of hotel. The hotel should include free parking and should include free wifi. Once the user find the hotel the user want to book it for 6 people and 3 nights starting from tuesday. If the booking fails how about 2 nights. Make sure the uer get the reference number. \textbackslash n\\
        Chat History & dialogue: \\
        &user: am looking for a place to to stay that has cheap price range it should be in a type of hotel. \textbackslash n\\ 
        &System: [missing sentence] \textbackslash n\\
        &user: no, i just need to make sure it's cheap. oh, and i need parking. \textbackslash n\\ 
        &System: [missing sentence] \textbackslash n\\
        Task & Replace all the "[missing sentence]" in the dialogue. please output the entire dialogue.\\
        \hline
            \end{tabular}
    }
    \label{tab:GPT3.5 missing sentence completion prompt example}
\end{table*}

\begin{table*}[t]
    \centering
    \small
    \caption{Structured user prompt example for Next Response Generation data augmentation.}
    \resizebox{\textwidth}{!}{%
    \begin{tabular}{lp{15cm}}
    \hline
        \textbf{Components} & \textbf{Prompt}\\
        \hline
          Task & Task: Generate the next user response according to the given goal and chat history. Your response must start with 'user:'! \textbackslash n\\
          Goal & Goal: The user is looking for a place to stay. The hotel should be in the cheap price range and should be in the type of hotel. The hotel should include free parking and should include free wifi. Once the user find the hotel the user want to book it for 6 people and 3 nights starting from tuesday. If the booking fails how about 2 nights. Make sure the uer get the reference number. \textbackslash n\\
         Chat History & Chat history: \\
         &user: am looking for a place to to stay that has cheap price range it should be in a type of hotel. \textbackslash n\\ 
         &system: Okay, do you have a specific area you want to stay in? \textbackslash n\\
        &user: no, i just need to make sure it's cheap. oh, and i need parking. \textbackslash n\\ 
        &system: I found 1 cheap hotel for you that includes parking. Do you like me to book it? \textbackslash n\\
        \hline
    \end{tabular}
    }
    \label{tab:Next Response Generation prompt example}
\end{table*}

\begin{table*}[t]
    \centering
    \small
    \caption{Structured user prompt example for Par. data augmentation.}
    \resizebox{\textwidth}{!}{%
    \begin{tabular}{llp{13.8cm}}
    \hline
    \textbf{Stages} &\textbf{Components} & \textbf{Prompt}\\
    \hline
     Stage 1 & Task Summary & A customer and a server line assistant are in dialogue. Replace each existing system response with a response you would have said if you were the system. Ensure the new responses logically follow the preceding dialogue and lead naturally into the unchanged user responses. The output should remain the same format as the dialogue!\textbackslash n\\
     & Dialogue & user: am looking for a place to to stay that has cheap price range it should be in a type of hotel. \textbackslash n\\ 
     &&system: Okay, do you have a specific area you want to stay in? \textbackslash n\\
     &&user: no, i just need to make sure it's cheap. oh, and i need parking. \textbackslash n\\ 
     &&system: I found 1 cheap hotel for you that includes parking. Do you like me to book it? \textbackslash n\\
     \hline
     Stage 2 & Dialogue & here is the chat history:\\
     && user: am looking for a place to to stay that has cheap price range it should be in a type of hotel. \textbackslash n\\ 
     &&system: Sure thing! Are you looking for a specific area or just anywhere with affordable prices? \textbackslash n\\
     &&user: no, i just need to make sure it's cheap. oh, and i need parking. \textbackslash n\\ 
     &&system: I've found a hotel that fits your criteria - it's budget-friendly and offers parking. Would you like me to proceed with the booking? \textbackslash n\\
     & Task Summary & please rewrite each sentence starting with 'user:'. do not copy and paste, you should rephrase it! you should add the exactly unchanged system response to form the output.\\
     \hline
    \end{tabular}
    }
    \label{tab: par. prompt example}
\end{table*}

\begin{table*}[t]
    \centering
    \small
    \caption{Structured user prompt example for G2D data augmentation part 1.}
    \resizebox{\textwidth}{!}{%
    \begin{tabular}{lp{12cm}}
    \hline
         \textbf{Components} & \textbf{Prompt}\\
        \hline
         Task Summary& Task: create a dialogue between a user and a system. \textbackslash n\\
         Example Dialogue& Here is a demonstration for the dialogue you are going to generate: \textbackslash n\\
         & user: Hello, I'm looking for a guesthouse on the westside to stay. \textbackslash n\\
         & system: I am sorry I have no guest houses in the west. Can I look one up in another area for you? \textbackslash n\\
         & user: Are there any other places like that that also have free parking and free wifi in that area? \textbackslash n\\
         & system: I have found one that is cheap with 4 stars, and one that is moderately priced with 3 stars? Do either of those sound appealing to you? \textbackslash n\\
         & user: The cheap one with the better rating sounds appealing. \textbackslash n\\
         & system: Great! That's Finches Bed and Breakfast. It's located at 144 Thornton Road cb30nd. Would you like to book a stay? \textbackslash n\\
         & user: Please book it for Friday,thank you. \textbackslash n\\
         & system: Can you please tell me how many people will be staying and for how many nights?
        \textbackslash n\\
        & user: Yes, 3 people for 5 nights starting on monday. \textbackslash n\\
        & system: Booking was unsuccessful. Want to try another hotel? \textbackslash n\\
        & user: Yes, in the same price range please. \textbackslash n\\
        & system: Worth House in the North is 4 stars and cheap. Would you like to try it? \textbackslash n\\
        & user: The amount of stars do not matter as long as it is in the same price range. Does Worth House in the North include free parking and internet? \textbackslash n\\
        & system: Okay well I was able to book you at The Worth House and your reference number is 86XVFEUY. \textbackslash n\\
        & user: Thanks, that's all I need. Have a nice day. \textbackslash n\\
        & system: You are welcome. Also for your reference the Worth House does have free parking and wifi. May I help you with anything else? \textbackslash n\\
        & user: No thanks. Thanks again for your help. \textbackslash n\\
        & system: Enjoy your stay! \textbackslash n\\
        Goal-Specific Instructions for User & For the dialog that you have to generate in this Section, the instructions for the "user" are the following: The user is looking for a place to stay. The hotel should be in the cheap price range and should be in the type of hotel. The hotel should include free parking and should include free wifi. Once the user find the hotel the user want to book it for 6 people and 3 nights starting from tuesday. If the booking fails how about 2 nights. Make sure the uer get the reference number. Every user message should be followed by a system message. Be polite and don’t forget to say goodbye. \\
        Slot and Domain Knowledge for System & For the dialog that you have to generate in this section, the instructions for the "system" are the following:\\
        & "internet": \{ "description": "whether the hotel has internet", "is\_categorical": true, "possible\_values": $\left[ \text{"free"}, \text{"no"}, \text{"yes"} \right]$ \}\\
        &"parking": \{ "description": "whether the hotel has parking", "is\_categorical": true, "possible\_values": $\left[ \text{"free"}, \text{"no"}, \text{"yes"} \right]$ \},\\
        & "area": \{ "description": "area or place of the hotel", "is\_categorical": true, "possible\_values": $\left[\text{"centre"}, \text{"east"}, \text{"north"}, \text{"south"}, \text{"west"} \right]$ \}\\
        & "stars": \{ "description": "star rating of the hotel", "is\_categorical": true, "possible\_values":  $\left[ \text{"0"}, \text{"1"}, \text{"2"}, \text{"3"}, \text{"4"}, \text{"5"} \right]$ \}\\
        & "price range": \{"description": "price budget of the hotel", "is\_categorical": true, "possible\_values": $\left[\text{"expensive"}, \text{"cheap"}, \text{"moderate"} \right]$ \}\\
        & "type": \{"description": "what is the type of the hotel", "is\_categorical": true, "possible\_values": $\left[\text{"guesthouse"}, \text{"hotel"} \right]$ \}\\
        & "name": \{ "description": "name of the hotel", "is\_categorical": false, "possible\_values": $\left[ \right]$ \}\\
        & "book people": \{ "description": "number of people for the hotel booking", "is\_categorical": false, "possible\_values": $\left[ \right]$ \}\\
        & "book stay": \{ "description": "length of stay at the hotel", "is\_categorical": false, "possible\_values": $\left[ \right]$ \}\\
        & "book day": \{ "description": "day of the hotel booking", "is\_categorical": true, "possible\_values": $\left[\text{"monday"}, \text{"tuesday"}, \text{"wednesday"}, \text{"thursday"}, \text{"friday"}, \text{"saturday"}, \text{"sunday"} \right]$ \}\\
        & "phone": \{ "description": "phone number of the hotel", "is\_categorical": false, "possible\_values": $\left[ \right]$ \}\\
        & "postcode": \{ "description": "postcode of the hotel", "is\_categorical": false, "possible\_values": $\left[ \right]$ \}\\
        & "address": \{ "description": "address of the hotel", "is\_categorical": false, "possible\_values": $\left[ \right]$ \}\\
        & "ref": \{ "description": "reference number of the hotel booking", "is\_categorical": false, "possible\_values": $\left[ \right]$ \}\\
        & "choice": \{ "description": "number of hotels that meet the requirement", "is\_categorical": false, "possible\_values": $\left[ \right]$ \}\\
        &Domain of knowledge needed (include everything is not mandatory): (parking, area, star rating, price range, type of hotel, name of hotel, book people number, length of stay, book day, phone number, postcode, address of hotel, reference number, number of hotel meet requirement) \\
         \hline
    \end{tabular}
    }
    \label{tab: G2D prompt example}
\end{table*}

\begin{table*}[t]
    \centering
    \small
    \caption{Structured user prompt example for G2D data augmentation part 2.}
    \resizebox{\textwidth}{!}{%
    \begin{tabular}{lp{12cm}}
    \hline
         \textbf{Components} & \textbf{Prompt}\\
        \hline       
        Conversation Termination Conditions & please generate a dialogue according to the instructions. If you achieve your goal, express your thanks and generate **\"[END]\"** token. If you think the assistant can not help you or the conversation falls into a infinite loop, generate **\"[STOP]\"** token.\\
        Sensitive Information Masking & Please mask the following information in the generated dialogue: (name of hotel as [hotel name], phone number as [phone number], postcode as [postcode], address of hotel as [address], reference number as [ref]). \\
         \hline
    \end{tabular}
    }
    \label{tab: G2D prompt example part 2}
\end{table*}

\begin{table*}[t]
    \centering
    \small
    \caption{Structured user prompt example for E2E Convo. data augmentation using GPT3.5.}
    \resizebox{\textwidth}{!}{%
    \begin{tabular}{p{4cm}p{12cm}}
    \hline
    \textbf{Components} & \textbf{Prompt}\\
    \hline
    Task Summary & Task: act as a user communicating with a system.\\
    Example Dialogue& Here is a demonstration for the dialogue you are going to generate: \textbackslash n\\
         & user: Hello, I'm looking for a guesthouse on the westside to stay. \textbackslash n\\
         & system: I am sorry I have no guest houses in the west. Can I look one up in another area for you? \textbackslash n\\
         & user: Are there any other places like that that also have free parking and free wifi in that area? \textbackslash n\\
         & system: I have found one that is cheap with 4 stars, and one that is moderately priced with 3 stars? Do either of those sound appealing to you? \textbackslash n\\
         & user: The cheap one with the better rating sounds appealing. \textbackslash n\\
         & system: Great! That's Finches Bed and Breakfast. It's located at 144 Thornton Road cb30nd. Would you like to book a stay? \textbackslash n\\
         & user: Please book it for Friday,thank you. \textbackslash n\\
         & system: Can you please tell me how many people will be staying and for how many nights?
        \textbackslash n\\
        & user: Yes, 3 people for 5 nights starting on monday. \textbackslash n\\
        & system: Booking was unsuccessful. Want to try another hotel? \textbackslash n\\
        & user: Yes, in the same price range please. \textbackslash n\\
        & system: Worth House in the North is 4 stars and cheap. Would you like to try it? \textbackslash n\\
        & user: The amount of stars do not matter as long as it is in the same price range. Does Worth House in the North include free parking and internet? \textbackslash n\\
        & system: Okay well I was able to book you at The Worth House and your reference number is 86XVFEUY. \textbackslash n\\
        & user: Thanks, that's all I need. Have a nice day. \textbackslash n\\
        & system: You are welcome. Also for your reference the Worth House does have free parking and wifi. May I help you with anything else? \textbackslash n\\
        & user: No thanks. Thanks again for your help. \textbackslash n\\
        & system: Enjoy your stay! \textbackslash n\\
        Role-Specific Instructions (User) & For the dialog that you have to generate in this Section, the instructions for the "user" are the following: The user is looking for a place to stay. The hotel should be in the cheap price range and should be in the type of hotel. The hotel should include free parking and should include free wifi. Once the user find the hotel the user want to book it for 6 people and 3 nights starting from tuesday. If the booking fails how about 2 nights. Make sure the user get the reference number. Every user message should be followed by a system message. Be polite and don’t forget to say goodbye. \textbackslash n\\
        Role-Specific Instructions (System) & I will be the system. \textbackslash n\\
        Conversation Termination Conditions & please generate a dialogue according to the goal. If you achieve your goal (booking sucessful or find the hotel), express your thanks and generate **\"[END]\"** token. If you think the assistant can not help you or the conversation falls into a infinite loop, generate **\"[STOP]\"** token. \textbackslash n\\
        Sensitive Information Masking& please mask the following information in the generated dialogue: (name of hotel as [hotel name], phone number as [phone number], postcode as [postcode], address of hotel as [address], reference number as [ref]). \textbackslash n The output user response should be in the format of "user:...". \textbackslash n It should be only one sentence.\\
    \hline
    \end{tabular}
    }
    \label{tab: E2E Convo. GPT3.5 user prompt example}
\end{table*}

\begin{table*}[t]
    \centering
    \small
    \caption{Structured system prompt example for E2E Convo. data augmentation using GPT3.5.}
    \resizebox{\textwidth}{!}{%
    \begin{tabular}{lp{12cm}}
    \hline
\textbf{Components} & \textbf{Prompt}\\
    \hline
        Task Summary & Task: act as a system communicating with a user.\\
    Example Dialogue& Here is a demonstration for the dialogue you are going to generate: \textbackslash n\\
         & user: Hello, I'm looking for a guesthouse on the westside to stay. \textbackslash n\\
         & system: I am sorry I have no guest houses in the west. Can I look one up in another area for you? \textbackslash n\\
         & user: Are there any other places like that that also have free parking and free wifi in that area? \textbackslash n\\
         & system: I have found one that is cheap with 4 stars, and one that is moderately priced with 3 stars? Do either of those sound appealing to you? \textbackslash n\\
         & user: The cheap one with the better rating sounds appealing. \textbackslash n\\
         & system: Great! That's Finches Bed and Breakfast. It's located at 144 Thornton Road cb30nd. Would you like to book a stay? \textbackslash n\\
         & user: Please book it for Friday,thank you. \textbackslash n\\
         & system: Can you please tell me how many people will be staying and for how many nights?
        \textbackslash n\\
        & user: Yes, 3 people for 5 nights starting on monday. \textbackslash n\\
        & system: Booking was unsuccessful. Want to try another hotel? \textbackslash n\\
        & user: Yes, in the same price range please. \textbackslash n\\
        & system: Worth House in the North is 4 stars and cheap. Would you like to try it? \textbackslash n\\
        & user: The amount of stars do not matter as long as it is in the same price range. Does Worth House in the North include free parking and internet? \textbackslash n\\
        & system: Okay well I was able to book you at The Worth House and your reference number is 86XVFEUY. \textbackslash n\\
        & user: Thanks, that's all I need. Have a nice day. \textbackslash n\\
        & system: You are welcome. Also for your reference the Worth House does have free parking and wifi. May I help you with anything else? \textbackslash n\\
        & user: No thanks. Thanks again for your help. \textbackslash n\\
        & system: Enjoy your stay! \textbackslash n\\
        Role-Specific Instructions (System) & for this dialogue, you are the system, here is the goal for system: Do not copy anything from the demonstration.please do not repeat yourself.Note that you should not make booking unless the goal explicitly mentioned a booking.you can only use the information provided in chat history.you can only generate one sentence each time. \textbackslash n\\
        system version Goal & The system needs to find a hotel in the cheap price range, with the type specified as "hotel." The hotel must offer free parking and free Wi-Fi. Once a suitable hotel is found, the system should proceed to book it for 6 people for 3 nights, starting from Tuesday. If the booking fails, the system should attempt to book for 2 nights. The system must ensure that the user receives a reference number for the booking.\\
        Role-Specific Instructions (User) & I will be the user. \textbackslash n\\
        Conversation Termination Conditions & please generate a dialogue according to the goal. If you achieve your goal (booking sucessful or find the hotel), express your thanks and generate **\"[END]\"** token. If you think the assistant can not help you or the conversation falls into a infinite loop, generate **\"[STOP]\"** token. \textbackslash n\\
        Sensitive Information Masking& please mask the following information in the generated dialogue: (name of hotel as [hotel name], phone number as [phone number], postcode as [postcode], address of hotel as [address], reference number as [ref]). \textbackslash n The output user response should be in the format of "user:...". \textbackslash n It should be only one sentence.\\
    \hline
    \end{tabular}
    }
    \label{tab: E2E Convo. GPT3.5 system prompt example}
\end{table*}

\begin{table*}[t]
    \centering
    \small
    \caption{Structured user prompt example for E2E Convo. data augmentation using Llama 70B part 1.}
    \resizebox{\textwidth}{!}{%
    \begin{tabular}{lp{12cm}}
    \hline
    \textbf{Components} & \textbf{Prompt}\\
    \hline
    Task Summary & Task: Simulate as an user with a particular goal and generate one response to a hotel service system. Response must start with "user:". After you achieved all your goals, end the conversation and generate "[END]" token. If you think the system cannot help you or the conversation falls into an infinite loop, generate a "[STOP]" token. The response must be one line only!\\
    Slot and Domain Knowledge for System & The information you can ask for or provide include:\\
            & "internet": \{ "description": "whether the hotel has internet", "is\_categorical": true, "possible\_values": $\left[ \text{"free"}, \text{"no"}, \text{"yes"} \right]$ \}\\
        &"parking": \{ "description": "whether the hotel has parking", "is\_categorical": true, "possible\_values": $\left[ \text{"free"}, \text{"no"}, \text{"yes"} \right]$ \},\\
        & "area": \{ "description": "area or place of the hotel", "is\_categorical": true, "possible\_values": $\left[\text{"centre"}, \text{"east"}, \text{"north"}, \text{"south"}, \text{"west"} \right]$ \}\\
        & "stars": \{ "description": "star rating of the hotel", "is\_categorical": true, "possible\_values":  $\left[ \text{"0"}, \text{"1"}, \text{"2"}, \text{"3"}, \text{"4"}, \text{"5"} \right]$ \}\\
        & "price range": \{"description": "price budget of the hotel", "is\_categorical": true, "possible\_values": $\left[\text{"expensive"}, \text{"cheap"}, \text{"moderate"} \right]$ \}\\
        & "type": \{"description": "what is the type of the hotel", "is\_categorical": true, "possible\_values": $\left[\text{"guesthouse"}, \text{"hotel"} \right]$ \}\\
        & "name": \{ "description": "name of the hotel", "is\_categorical": false, "possible\_values": $\left[ \right]$ \}\\
        & "book people": \{ "description": "number of people for the hotel booking", "is\_categorical": false, "possible\_values": $\left[ \right]$ \}\\
        & "book stay": \{ "description": "length of stay at the hotel", "is\_categorical": false, "possible\_values": $\left[ \right]$ \}\\
        & "book day": \{ "description": "day of the hotel booking", "is\_categorical": true, "possible\_values": $\left[\text{"monday"}, \text{"tuesday"}, \text{"wednesday"}, \text{"thursday"}, \text{"friday"}, \text{"saturday"}, \text{"sunday"} \right]$ \}\\
        & "phone": \{ "description": "phone number of the hotel", "is\_categorical": false, "possible\_values": $\left[ \right]$ \}\\
        & "postcode": \{ "description": "postcode of the hotel", "is\_categorical": false, "possible\_values": $\left[ \right]$ \}\\
        & "address": \{ "description": "address of the hotel", "is\_categorical": false, "possible\_values": $\left[ \right]$ \}\\
        & "ref": \{ "description": "reference number of the hotel booking", "is\_categorical": false, "possible\_values": $\left[ \right]$ \}\\
        & "choice": \{ "description": "number of hotels that meet the requirement", "is\_categorical": false, "possible\_values": $\left[ \right]$ \}\\
        & Information with “mask\_token” specified must be replaced by corresponding token in your response, unless it is provided by the system or in your goal. Do not ask for or provide other information. You do not need to confirm details with the system unless it is ambiguous.\\
        Example Dialogue & Here is a demonstration partial dialogue unrelated to your own goal:\textbackslash n\\
         & user: Hello, I'm looking for a guesthouse on the westside to stay. \textbackslash n\\
         & system: I am sorry I have no guest houses in the west. Can I look one up in another area for you? \textbackslash n\\
         & user: Are there any other places like that that also have free parking and free wifi in that area? \textbackslash n\\
         & system: I have found one that is cheap with 4 stars, and one that is moderately priced with 3 stars? Do either of those sound appealing to you? \textbackslash n\\
         & user: The cheap one with the better rating sounds appealing. \textbackslash n\\
         & system: Great! That's Finches Bed and Breakfast. It's located at 144 Thornton Road cb30nd. Would you like to book a stay? \textbackslash n\\
         & user: Please book it for Friday,thank you. \textbackslash n\\
         & system: Can you please tell me how many people will be staying and for how many nights?
        \textbackslash n\\
        & user: Yes, 3 people for 5 nights starting on monday. \textbackslash n\\
        & system: Booking was unsuccessful. Want to try another hotel? \textbackslash n\\
        & user: Yes, in the same price range please. \textbackslash n\\
        & system: Worth House in the North is 4 stars and cheap. Would you like to try it? \textbackslash n\\
        & user: The amount of stars do not matter as long as it is in the same price range. Does Worth House in the North include free parking and internet? \textbackslash n\\
        & system: Okay well I was able to book you at The Worth House and your reference number is 86XVFEUY. \textbackslash n\\
        & user: Thanks, that's all I need. Have a nice day. \textbackslash n\\
        & system: You are welcome. Also for your reference the Worth House does have free parking and wifi. May I help you with anything else? \textbackslash n\\
        & user: No thanks. Thanks again for your help. \textbackslash n\\
        & system: Enjoy your stay! \textbackslash n\\
        & Do not copy anything from the demonstration! \textbackslash n\\
    \hline
    \end{tabular}
    }
    \label{tab: E2E Convo. Llama 70B user prompt example}
\end{table*}

\begin{table*}[t]
    \centering
    \small
    \caption{Structured user prompt example for E2E Convo. data augmentation using Llama 70B part 2.}
    \resizebox{\textwidth}{!}{%
    \begin{tabular}{lp{12cm}}
    \hline
    \textbf{Components} & \textbf{Prompt}\\
    \hline
        Role-Specific Instructions (User) & Here is your goal: \textbackslash n\\
        & The user is looking for a place to stay. The hotel should be in the cheap price range and should be in the type of hotel. The hotel should include free parking and should include free wifi. Once the user find the hotel the user want to book it for 6 people and 3 nights starting from tuesday. If the booking fails how about 2 nights. Make sure the user get the reference number.\\
        & Note that you should not make booking unless the goal explicitly mentioned a booking. Do not ask for or provide information not specified in the goal. If you are looking for a specific hotel which cannot be found, and the goal does not specify alternative action, end the conversation.\\
    \hline
    \end{tabular}
    }
    \label{tab: E2E Convo. Llama 70B user prompt example part 2}
\end{table*}

\begin{table*}[t]
    \centering
    \small
    \caption{Structured prompt system example for E2E Convo. data augmentation using Llama 70B.}
    \resizebox{\textwidth}{!}{%
    \begin{tabular}{lp{12cm}}
    \hline
    \textbf{Components} & \textbf{Prompt}\\
    \hline
    Task Summary & Task: Simulate as a hotel service system and generate one response to a user. Response must start with "system:". If and only if the user has no more queries or generated "[END]", end the conversation and generate "[END]" token. If you think the conversation falls into an infinite loop, generate a "[STOP]" token.\\
    Slot and Domain Knowledge for System & The information you can ask for or provide include:\\
        & "internet": \{ "description": "whether the hotel has internet", "is\_categorical": true, "possible\_values": $\left[ \text{"free"}, \text{"no"}, \text{"yes"} \right]$ \}\\
        &"parking": \{ "description": "whether the hotel has parking", "is\_categorical": true, "possible\_values": $\left[ \text{"free"}, \text{"no"}, \text{"yes"} \right]$ \},\\
        & "area": \{ "description": "area or place of the hotel", "is\_categorical": true, "possible\_values": $\left[\text{"centre"}, \text{"east"}, \text{"north"}, \text{"south"}, \text{"west"} \right]$ \}\\
        & "stars": \{ "description": "star rating of the hotel", "is\_categorical": true, "possible\_values":  $\left[ \text{"0"}, \text{"1"}, \text{"2"}, \text{"3"}, \text{"4"}, \text{"5"} \right]$ \}\\
        & "price range": \{"description": "price budget of the hotel", "is\_categorical": true, "possible\_values": $\left[\text{"expensive"}, \text{"cheap"}, \text{"moderate"} \right]$ \}\\
        & "type": \{"description": "what is the type of the hotel", "is\_categorical": true, "possible\_values": $\left[\text{"guesthouse"}, \text{"hotel"} \right]$ \}\\
        & "name": \{ "description": "name of the hotel", "is\_categorical": false, "possible\_values": $\left[ \right]$ \}\\
        & "book people": \{ "description": "number of people for the hotel booking", "is\_categorical": false, "possible\_values": $\left[ \right]$ \}\\
        & "book stay": \{ "description": "length of stay at the hotel", "is\_categorical": false, "possible\_values": $\left[ \right]$ \}\\
        & "book day": \{ "description": "day of the hotel booking", "is\_categorical": true, "possible\_values": $\left[\text{"monday"}, \text{"tuesday"}, \text{"wednesday"}, \text{"thursday"}, \text{"friday"}, \text{"saturday"}, \text{"sunday"} \right]$ \}\\
        & "phone": \{ "description": "phone number of the hotel", "is\_categorical": false, "possible\_values": $\left[ \right]$ \}\\
        & "postcode": \{ "description": "postcode of the hotel", "is\_categorical": false, "possible\_values": $\left[ \right]$ \}\\
        & "address": \{ "description": "address of the hotel", "is\_categorical": false, "possible\_values": $\left[ \right]$ \}\\
        & "ref": \{ "description": "reference number of the hotel booking", "is\_categorical": false, "possible\_values": $\left[ \right]$ \}\\
        & "choice": \{ "description": "number of hotels that meet the requirement", "is\_categorical": false, "possible\_values": $\left[ \right]$ \}\\
        & Information with “mask\_token” specified must be replaced by corresponding token in your response, unless it is provided by the system or in your goal. Do not ask for or provide other information. You do not need to confirm details with the system unless it is ambiguous.\\
        Example Dialogue & Here is a demonstration partial dialogue unrelated to your own goal:\textbackslash n\\
         & user: Hello, I'm looking for a guesthouse on the westside to stay. \textbackslash n\\
         & system: I am sorry I have no guest houses in the west. Can I look one up in another area for you? \textbackslash n\\
         & user: Are there any other places like that that also have free parking and free wifi in that area? \textbackslash n\\
         & system: I have found one that is cheap with 4 stars, and one that is moderately priced with 3 stars? Do either of those sound appealing to you? \textbackslash n\\
         & user: The cheap one with the better rating sounds appealing. \textbackslash n\\
         & system: Great! That's Finches Bed and Breakfast. It's located at 144 Thornton Road cb30nd. Would you like to book a stay? \textbackslash n\\
         & user: Please book it for Friday,thank you. \textbackslash n\\
         & system: Can you please tell me how many people will be staying and for how many nights?
        \textbackslash n\\
        & user: Yes, 3 people for 5 nights starting on monday. \textbackslash n\\
        & system: Booking was unsuccessful. Want to try another hotel? \textbackslash n\\
        & user: Yes, in the same price range please. \textbackslash n\\
        & system: Worth House in the North is 4 stars and cheap. Would you like to try it? \textbackslash n\\
        & user: The amount of stars do not matter as long as it is in the same price range. Does Worth House in the North include free parking and internet? \textbackslash n\\
        & system: Okay well I was able to book you at The Worth House and your reference number is 86XVFEUY. \textbackslash n\\
        & user: Thanks, that's all I need. Have a nice day. \textbackslash n\\
        & system: You are welcome. Also for your reference the Worth House does have free parking and wifi. May I help you with anything else? \textbackslash n\\
        & user: No thanks. Thanks again for your help. \textbackslash n\\
        & system: Enjoy your stay! \textbackslash n\\
        & Do not copy anything from the demonstration! \textbackslash n\\
        \hline
    \end{tabular}
    }
    \label{tab: E2E Convo. Llama 70B system prompt example}
\end{table*}

\end{document}